\documentclass[10pt,twocolumn,letterpaper]{article}

\usepackage[pagenumbers]{cvpr} % To force page numbers, e.g. for an arXiv version

\newcommand{\TODO}[1]{\textbf{\color{red}[TODO: #1]}}
\usepackage{eccvabbrv}
\usepackage{soul}
\usepackage{float}
\usepackage{pifont}
\usepackage{listings}
\usepackage{array}
\usepackage{colortbl}
\usepackage{multirow}
\usepackage{wrapfig}
\usepackage{placeins}
\usepackage{fontawesome5}
\usepackage[accsupp]{axessibility}

\definecolor{qualitative_green}{RGB}{0,176,80}
\definecolor{qualitative_orange}{RGB}{255,192,0}
\definecolor{qualitative_blue}{RGB}{68,114,196}
\definecolor{codegreen}{rgb}{0,0.6,0}
\definecolor{negativeorange}{RGB}{214,39,40}
\definecolor{LightBlue}{rgb}{0.68,0.85,0.9}
\definecolor{Srijan}{rgb}{0.8,0,0.8}

\newcommand{\xmark}{\ding{55}}%
\newcommand{\train}{\textcolor{red}{\faFire}}
\newcommand{\frozen}{\textcolor{blue}{\faSnowflake}}
\newcommand{\methodname}{\textsc{VisCoP\xspace}}
\newcommand{\methodexpansion}{\underline{\textbf{Vis}}ion \underline{\textbf{Co}}ntextualized \underline{\textbf{P}}robing}
\newcommand{\Srijan}[1]{\textcolor{Srijan}{#1}}

\definecolor{cvprblue}{rgb}{0.21,0.49,0.74}
\usepackage[pagebackref,breaklinks,colorlinks,allcolors=cvprblue]{hyperref}

\def\paperID{*****} % *** Enter the Paper ID here
\def\confName{CVPR}
\def\confYear{2026}

\title{
    \begin{minipage}{0.125\textwidth}
        \includegraphics[width=0.85\textwidth]{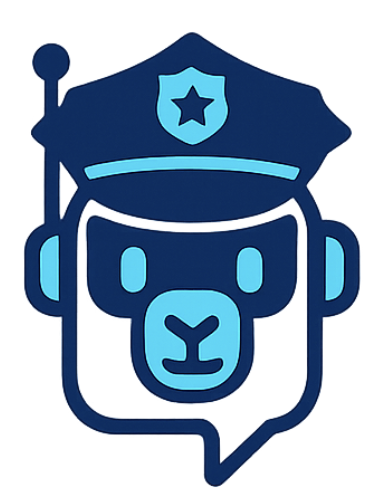}
    \end{minipage}
    \hspace{-0.18\textwidth}
    \begin{minipage}{0.9\textwidth}
        \centering
        \methodname: Visual Probing for Video Domain \\ Adaptation of Vision Language Models
    \end{minipage}
}

\author{\textbf{Dominick Reilly}$^{1}$ \quad \textbf{Manish Kumar Govind}$^{1}$ \quad \textbf{Le Xue}$^{2}$ \quad \textbf{Srijan Das}$^{1}$ \vspace{0.1cm} \\
$^{1}$University of North Carolina at Charlotte \quad $^{2}$Elorian AI\vspace{0.1cm} \\
{\tt\small dreilly1@charlotte.edu}\\
\url{https://github.com/dominickrei/VisCoP}
}

\begin{document}
\maketitle
\begin{abstract}
  Large Vision Language Models (VLMs) excel at general visual reasoning tasks, but their performance degrades sharply when deployed in novel domains with substantial distribution shifts compared to what was seen during pretraining. Existing approaches to adapt VLMs to novel target domains rely on finetuning standard VLM components. Depending on which components are finetuned, these approaches either limit the VLM's ability to learn domain-specific features, or lead to  catastrophic forgetting of pre-existing capabilities.
    To address this, we introduce \methodexpansion~(\textbf{\methodname}), which augments the VLM's vision encoder with a compact set of learnable \textit{visual probes}, enabling domain-specific features to be learned with only minimal updates to the pretrained VLM components.
    We evaluate \methodname\ across three challenging domain adaptation scenarios: cross-view (exocentric $\rightarrow$ egocentric), cross-modal (RGB $\rightarrow$ depth), and cross-task (human understanding $\rightarrow$ robot control). Our experiments demonstrate that \methodname\ consistently outperforms existing domain adaptation strategies, achieving superior performance on the target domain, while better retaining capabilities from the source domain.
    % We release all code, models, and evaluation protocols: \url{https://github.com/dominickrei/VisCoP}.
\end{abstract}

\begin{figure}[t]
    \centering
    \includegraphics[width=\linewidth]{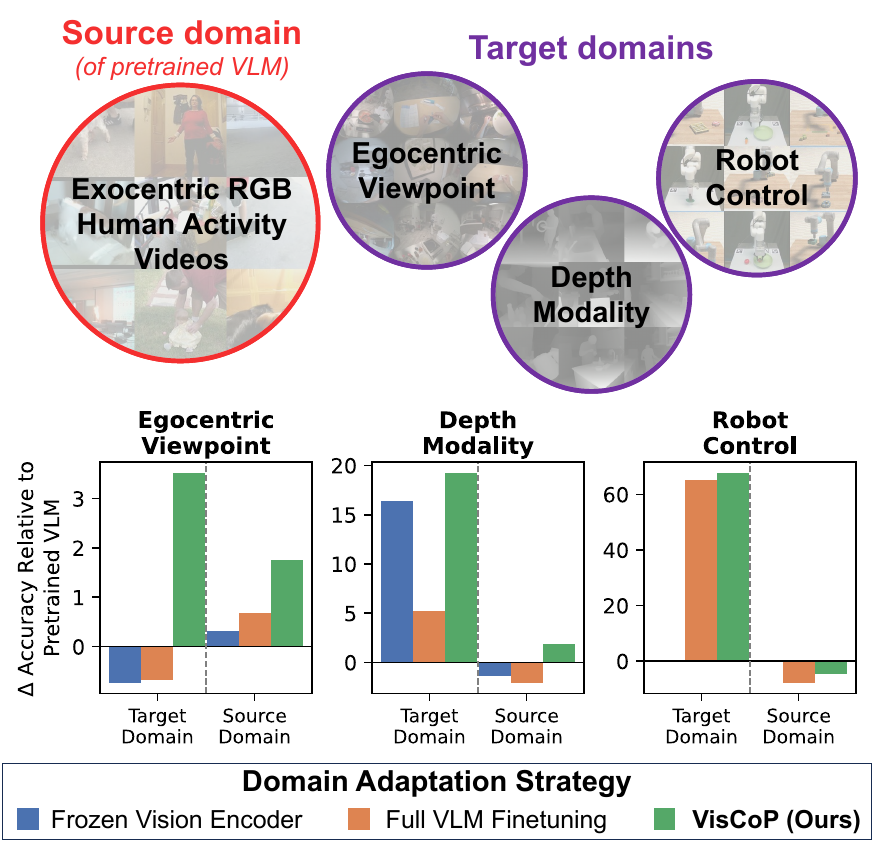}
    \caption{\textbf{Domain adaptation performance of different adaptation strategies.} \methodname\ achieves superior target domain performance while better retaining source domain knowledge compared to other strategies.}
    \vspace{-0.75cm}
    \label{fig:intro_domain_performance}
\end{figure}

\vspace{-0.5cm}
\section{Introduction}

Large Vision-Language Models (VLMs)~\cite{openai2025thinkwithimages,Qwen2.5-VL,damonlpsg2025videollama3,xue2025blip3} have demonstrated strong performance across diverse multimodal tasks, including open-ended video question answering~\cite{socratic,mmbench} and spatial reasoning~\cite{lai2023lisa,Ranasinghe2024LearningTL}.
These models couple pretrained vision encoders~\cite{radford2021openaiclip,zhai2023siglipv1} with Large Language Models (LLMs)~\cite{qwen2025qwen25technicalreport,meta2024llama3herdmodels} and are trained on large-scale, web-curated image-video-text corpora containing broad but generic visual concepts~\cite{zhang2024llavavideo,chen2024sharegpt4video,maaz2024videogptplus,rawal2024cinepile}.
However, when deployed in domains that exhibit a pronounced \emph{visual gap} arising from differences in viewpoint, sensing modality, or task structure, their performance degrades substantially under distribution shift.
This challenge is prevalent in video domain adaptation settings such as exocentric $\rightarrow$ egocentric viewpoint, RGB $\rightarrow$ depth modality, and visual understanding $\rightarrow$ robotic control, where learning domain-specific \emph{visual} representations is essential.
Crucially, VLMs must \emph{adapt while preserving} the broad knowledge acquired during pretraining: for example, a model that is adapted to egocentric video data should remain competent on exocentric video benchmarks.

A common strategy to address such distributional shift is partial fine-tuning of a pretrained VLM on domain-specific video instruction data.
Freezing the vision encoder and updating only lightweight components (e.g., the vision–language connector) preserves pretrained knowledge but restricts visual specialization.
In contrast, finetuning the vision encoder enables specialized visual understanding, albeit at the cost of catastrophic forgetting of pretrained knowledge~\cite{yang2023aim,zang2023overcoming,li2024peftclip}.
This trade-off is particularly severe when the dominant shift is visual.
Moreover, existing approaches rely primarily on the final-layer representations of a frozen vision encoder, discarding intermediate features that encode multi-level visual structure~\cite{li2023llavamed}.
We argue that such abstracted representations are insufficient for visually prominent domain shifts and pose the following question: \emph{how can a VLM leverage domain-specific visual signals across the depth of a pretrained vision encoder while avoiding catastrophic forgetting?}

To this end, we introduce \underline{\textbf{Vis}}ion \underline{\textbf{Co}}ntextualized \underline{\textbf{P}}robing (\textbf{\methodname}), a lightweight adaptation mechanism that enables pretrained VLMs to specialize to a target domain while preserving their general-purpose visual knowledge.
Motivated by the progressive emergence of semantic representations across transformer depths~\cite{vaswani2017attention,timesformer,liu2021videoswin}, \methodname~performs two way probing: it first extracts domain relevant signals from intermediate layers of a frozen vision encoder, and subsequently conditions the language model embedding space to enhance domain specific visual reasoning.
This is achieved through a compact set of learnable probe tokens that interact layer wise with intermediate visual representations, forming an alternative parameterized pathway for adaptation.
% Rather than adapting the backbone encoder itself, VisCoP learns a parallel representation pathway that extracts domain-specific signals from intermediate encoder features while preserving pretrained representations.
% Dr. Das: By leveraging multi level abstractions across the encoder depth, \methodname~captures domain specific visual patterns while mitigating catastrophic forgetting.
% By leveraging multi level abstractions across the encoder depth, \methodname~learns a parallel representation pathway that extracts domain-specific signals from intermediate encoder features while preserving pretrained representations, enabling the learning of domain specific visual patterns while mitigating catastrophic forgetting.
% In contrast to prior model centric approaches~\cite{blip2,flamingo,ha2024domainadaptqformer,ryoo2025xgenmmvidblip3videoneed32} and prompt tuning methods~\cite{jia2022visualprompttuning}, which primarily rely on final layer visual features or do not explicitly learn new visual cues within a frozen encoder, \methodname~enables multi layer probing that surfaces domain relevant representations that would otherwise be discarded.
%
By leveraging multi-level abstractions across the encoder depth, \methodname~learns a parallel representation pathway that extracts domain-specific signals from intermediate encoder features while preserving pretrained representations. This enables the model to learn domain-specific visual patterns while mitigating catastrophic forgetting.
In contrast, existing approaches~\cite{blip2,flamingo,ha2024domainadaptqformer,ryoo2025xgenmmvidblip3videoneed32,jia2022visualprompttuning} typically operate on the final-layer representations of the vision encoder or introduce prompt tokens that condition downstream reasoning without explicitly learning new visual cues \textit{within} the vision encoder.
This design provides a simple yet effective approach to video domain adaptation under prominent visual gaps without modifying the pretrained visual encoder parameters, which is not enabled by existing approaches.
Empirically, we find that the representations learned via the \methodname~adaptation pathway enable effective cross-view, cross-modal, and cross-task adaptation of VLMs, while retaining their broad capabilities learned during pretraining.
Metaphorically, the name \methodname\ reflects its role as a “\textit{traffic cop}”,
directing gradient flows away from the visual encoder and towards an alternative pathway for learning domain-specific visual features, avoiding the ``\textit{crash}" (catastrophic forgetting) that would otherwise occur if gradients flowed directly through the vision encoder.

\noindent To summarize, our key contributions:
\begin{enumerate}
\item We propose \methodname~(\methodexpansion), a novel video domain adaptation strategy for VLMs that learns domain-specific visual representations through layer-wise probing of a frozen vision encoder, enabling effective domain transfer and preventing catastrophic forgetting of multi-modal capabilities acquired during pretraining.

\item We establish a comprehensive evaluation setting for domain adaptation in VLMs, spanning three challenging target domains: cross-view (exocentric $\to$ egocentric), cross-modality (RGB $\to$ depth), and cross-task (action understanding $\to$ robotic control), along with standardized metrics to evaluate performance.
% We will release code and data to facilitate future research on domain adaptation in VLMs.
\item Our experiments demonstrate that VLMs trained with \methodname~outperform alternative domain adaptation strategies across diverse target domains, while retaining more knowledge of the source domain illustrated in Figure \ref{fig:intro_domain_performance}.
\end{enumerate}

\section{Related Works}

\noindent\textbf{Domain adaptation in vision-language encoders.}\quad
Domain adaptation of contrastively trained vision-language encoders, such as CLIP~\cite{radford2021openaiclip, vificlip}, is typically achieved through prompt tuning or adapter-based approaches.
Both strategies aim to learn domain-specific features while keeping the pretrained vision and text encoders frozen. To accomplish this, prompt tuning approaches~\cite{zhou2022coop, zhu2023prograd, yao2023kgcoop} introduce learnable prompt vectors as additional input to the text encoder, steering the model toward target domain. Adapter-based approaches~\cite{yang2023aim, gao2023clipadapter} insert lightweight trainable modules directly into the encoder space, thus updating their pretrained representations.
% Prompt tuning approaches~\cite{zhou2022coop, zhu2023prograd, yao2023kgcoop} leave the pretrained vision and text encoders frozen, and introduce learnable prompt vectors as additional context into the text encoder to ``steer" the model towards the target domain.
% Adapter-based approaches~\cite{yang2023aim, gao2023clipadapter} also freeze the pretrained encoders, but introduce trainable modules directly into the encoder space. For example, AIM~\cite{yang2023aim} inserts feed-forward layers between intermediate layers of the vision encoder, enabling the model to learn domain-specific features at the cost of modifying the representations of the ision encoder, which can lead to catastrophic forgetting of source domain knowledge.
Head2Toe~\cite{evci2022head2toe} similarly exploits intermediate representations of a frozen backbone, but passively selects static activations to train a linear classifier per target task, with no focus on source-domain retention.
In contrast to these approaches, \methodname~addresses the setting of domain adaptation in generative VLMs, enabling them to learn domain-specific features without requiring updates to the pretrained encoder representations.

\noindent\textbf{Domain adaptation in VLMs.}\quad
Domain adaptation in VLMs has largely been achieved through data-centric strategies rather than through architectural changes~\cite{cheng2025domainadaptationvlms}. Existing approaches typically leverage automated pipelines~\cite{mohbat2024llavachef, reilly2025llavidal} or closed-source VLMs~\cite{li2023llavamed, chen2024huatuogpt_medicaladaptvlm} to curate visual-instruction pairs from existing datasets in the target domain.
Their adaptation strategy usually follows a multi-stage training scheme similar to LLaVA~\cite{liu2023_llava}, where different VLM components are selectively trained at each stage.
However, the choice of trainable components creates a trade-off between extracting domain-specific features and retaining pretrained knowledge.
Training only lightweight connectors retains pretrained knowledge but limits domain-specific visual understanding, while training the vision encoder enables specialized visual understanding at the cost of catastrophic forgetting.
\methodname~avoids this trade-off through the introduction of visual probes that extract domain-specific features from a frozen vision encoder, enabling adaptation without disrupting the pretrained visual representations.

% \noindent\textbf{Visual probing vs. visual compression.}\quad
% Several approaches employ learnable tokens to bridge vision and language modalities~\cite{ha2024domainadaptqformer, ryoo2025xgenmmvidblip3videoneed32, zohar2025apollo} through architectures leveraging the Q-Former and Perceiver Resampler modules.
% Q-Former~\cite{blip2} leverages learnable queries that cross-attend to representations from the final layer of the vision encoder, aggregating visual information into a reduced set of tokens for computational efficiency.
% Perceiver Resampler~\cite{flamingo} operates similarly, aiming to compress the visual representations into a fixed number of learnable tokens.
% The visual probes proposed in \methodname~differ fundamentally, as they are designed to \emph{extract} novel domain-specific visual representations rather than to simply \emph{compress} pretrained ones.
% This is enabled by their interaction with intermediate representations of the vision encoder, allowing the probes to extract domain-specific representations that are not propagated to the final representation of the pretrained vision encoder~\cite{radford2021openaiclip, zhai2023siglipv1}.

\noindent\textbf{Visual probing vs. visual compression and prompt tuning.}\quad
Several approaches employ learnable tokens to bridge vision and language modalities~\cite{ha2024domainadaptqformer, ryoo2025xgenmmvidblip3videoneed32, zohar2025apollo} through architectures leveraging the Q-Former, Perceiver Resampler, or prompt tuning mechanisms.
Q-Former~\cite{blip2} leverages learnable queries that cross-attend to representations from the final layer of the vision encoder, aggregating visual information into a reduced set of tokens for computational efficiency.
Perceiver Resampler~\cite{flamingo} operates similarly, aiming to compress the visual representations into a fixed number of learnable tokens.
Prompt tuning methods~\cite{li2021prefixtuning, zhou2022coop, jia2022visualprompttuning} introduce learnable tokens to steer downstream reasoning, but operate solely on fixed encoder representations without enabling learning of new visual features.
The visual probes proposed in \methodname~differ fundamentally, as they are designed to \emph{extract and learn} domain-specific visual representations rather than to simply \emph{compress} or \emph{condition on} pretrained ones.
This is enabled by their interaction with intermediate representations of the vision encoder, allowing the probes to extract domain-specific representations that are not propagated to the final representation of the pretrained vision encoder~\cite{radford2021openaiclip, zhai2023siglipv1}.

\section{Problem Formulation}

Let $\mathcal{S}$ denote the \emph{source domain}, on which the vision-language model $f_{\theta^0}$ has been pretrained, and let $\mathcal{T}$ denote the \emph{target domain}, the domain of interest for adaptation. The two domains differ in their underlying distributions (e.g., viewpoint, modality, or task), which causes $f_{\theta^0}$ to perform poorly when directly applied to $\mathcal{T}$.

Training supervision in these domains is provided as video-QA pairs $(v,q,a)$, where $v$ is a video, $q$ is an instruction or question, and $a$ is the corresponding response. While $f_{\theta^0}$ has been pretrained on samples $(v,q,a)\sim\mathcal{S}$, at adaptation time we only assume availability of target domain samples $(v,q,a)\sim\mathcal{T}$. The objective of domain adaptation is to update the pretrained parameters $\theta^0$ to obtain $\theta^\star$ that improves performance on domain $\mathcal{T}$, while retaining performance on domain $\mathcal{S}$. Formally,
\setlength{\abovedisplayskip}{3pt}
\setlength{\belowdisplayskip}{3pt}
\[
R_{\mathcal{T}}(\theta^\star) < R_{\mathcal{T}}(\theta^0)
\quad \text{and} \quad
R_{\mathcal{S}}(\theta^\star) \approx R_{\mathcal{S}}(\theta^0)
\]
where $R_{\mathcal{D}}$ denotes the VLM's expected autoregressive next-token prediction loss under domain $\mathcal{D}$.
In summary, our problem statement considers adaptation of a pretrained VLM to a novel domain using only video-QA pairs from that domain. The objective is to improve target-domain performance while minimizing catastrophic forgetting of source-domain capabilities. In the next section, we introduce our proposed method, which enables balanced domain adaptation under these constraints.

\section{Method: Video Domain-adaptive VLM}

Given a video input $\mathbf{V} = \{\boldsymbol{I}_t\}_{t=1}^{T}$ consisting of $T$ frames, the goal of the VLM is to generate the response corresponding to the input instruction in an autoregressive manner.

\subsection{Preliminary}

Existing VLMs for video representation learning~\cite{damonlpsg2025videollama3, Qwen2.5-VL} consist of three standard components: \textbf{(i)} a vision encoder that maps visual inputs into a sequence of spatio-temporal tokens, \textbf{(ii)} a vision-language connector that projects the visual tokens to the embedding space of a language model, and \textbf{(iii)} an LLM that processes the projected visual tokens jointly with language tokens to enable multi-modal reasoning.
For the input video $\mathbf{V}$, each frame $\boldsymbol{I}_t$ is processed independently by the vision encoder through a stack of $L$ transformer layers. The visual tokens after the $\ell$-th layer are denoted as
\[
\mathbf{X}_t^{\ell} \in \mathbb{R}^{N \times d_v}, \quad \ell = 1, \dots, L
\]
where $N$ is the number of spatial patch tokens per frame and $d_v$ is the embedding dimension of the vision encoder. Concatenating these tokens over time yields
$\mathbf{X}^{\ell} \in \mathbb{R}^{(TN) \times d_v}$
which represents the sequence of spatio-temporal visual tokens at the $\ell$-th layer of the vision encoder. The final layer outputs $\mathbf{X}^{L}$ are then projected to the language embedding space via a vision-language connector $\mathcal{C}$ to obtain the visual embeddings used as input to the LLM
\[
\mathbf{E} = \mathcal{C}(\mathbf{X}^{L}) \in \mathbb{R}^{(T\tilde{N}) \times d_{\text{lm}}}
\]
where $\tilde{N}$ is the number of visual tokens input to the LLM after spatial downsampling~\cite{damonlpsg2025videollama3}. and $d_{\text{lm}}$ is the embedding dimension of the LLM.

The VLM is then trained to optimize a standard autoregressive next token prediction loss. Specifically, given the visual embeddings $\mathbf{E}$ and the tokenized QA pair $(\mathbf{Q}, \mathbf{A})$, we optimize the likelihood of predicting $\mathbf{A}$ conditioned on the visual embeddings and the question
\[
P(\mathbf{A} \mid \mathbf{E}, \mathbf{Q}) =
\prod_{j=1}^{\textrm{Len}} P_{\boldsymbol{\theta}}(\mathbf{a}_j \mid \mathbf{E}, \mathbf{Q}, \mathbf{A}_{<j})
\]
where $\boldsymbol{\theta}$ are the trainable parameters of the VLM, $\textrm{Len}$ indicates the token length of $\mathbf{A}$, and $\mathbf{A}_{<j}$ represents the subsequence of answer tokens preceding position $j$.

For domain-adaptive post training of VLMs, finetuning the vision encoder of a pretrained VLM for a target domain $\mathcal{T}$ often leads to overfitting on $\mathcal{T}$ and catastrophic forgetting of the source domain~\cite{yang2023aim, zang2023overcoming, li2024peftclip}. To mitigate this trade-off, a domain-adaptive pathway is required that adapts the VLM to $\mathcal{T}$ while retaining performance on $\mathcal{S}$.
%To mitigate this trade-off, we propose \methodname, a video domain-adaptive representation learning approach that specializes to $\mathbf{T}$ while preserving source-generalizable behavior.

%we leave the vision encoder frozen, as its large capacity and extensive pretraining make it particularly susceptible to forgetting previously acquired knowledge.

\begin{figure*}[t!]
    \centering
    \includegraphics[width=\linewidth]{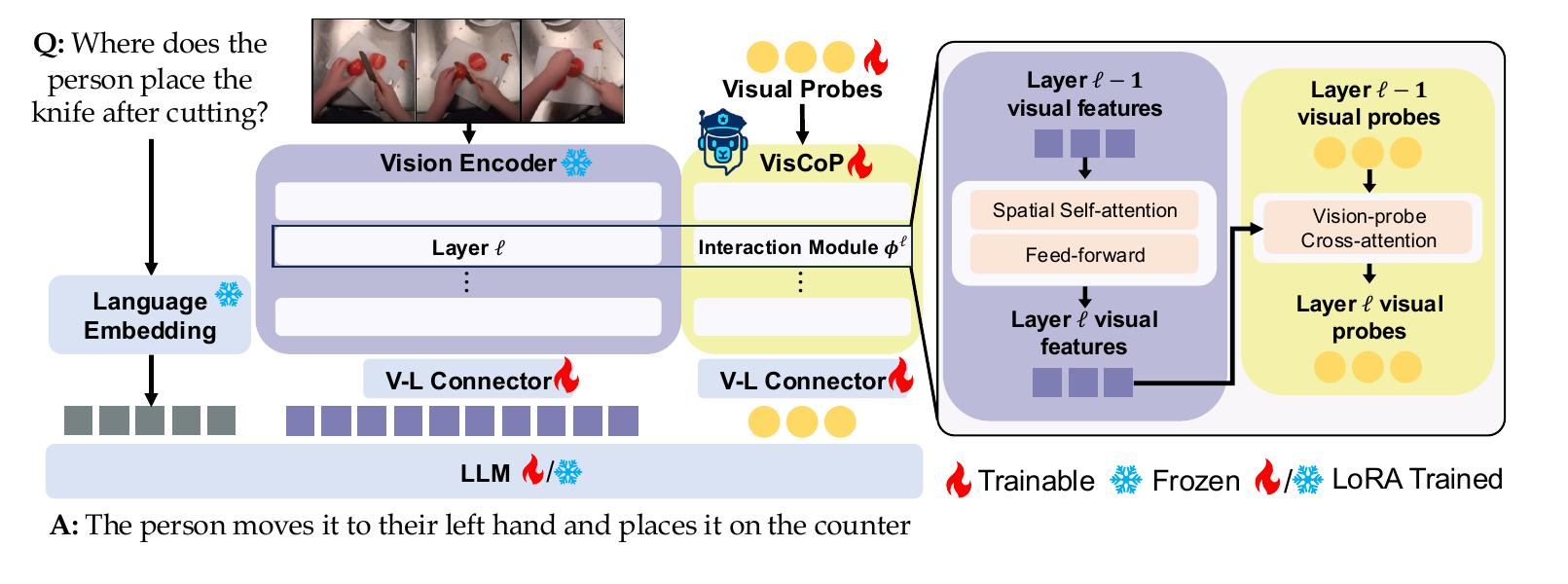}
    \vspace{-1cm}
    \caption{\textbf{Architecture of our proposed \methodname.} Learnable visual probes are conditioned on intermediate representations of a frozen vision encoder through vision-probe cross-attention, which extracts domain-specific features that may have otherwise been discarded by the frozen encoder.}
    \label{fig:method_diagram}
    \vspace{-0.5cm}
\end{figure*}

\subsection{\methodname: \underline{Vis}ion \underline{Co}ntextualized \underline{P}robing}

%Our goal is to train domain-specialized VLMs that avoid catastrophic forgetting without sacrificing their ability to learn domain-specific representations. However, directly finetuning large pretrained components often leads to catastrophic forgetting and overfitting to the target domain~\cite{yang2023aim, zang2023overcoming, li2024clipvlm_peft}. To mitigate this risk, we leave the vision encoder frozen, as its large capacity and extensive pretraining make it particularly susceptible to forgetting previously acquired knowledge.
% Directly updating large pretrained components often leads to overfitting and degraded generalization~\cite{yang2023aim, zang2023overcoming, li2024clipvlm_peft}. Thus, we choose to leave the vision encoder frozen during specialization, as its large capacity and relatively training instructions make it especially prone to overfitting.
To capture the relevant visual context that would otherwise be lost by freezing the vision encoder, we propose \methodexpansion~(\textbf{\methodname}), a mechanism that augments the vision encoder with a compact set of learnable tokens, called \textit{visual probes}, and an interaction module that acts as a semantic interface between the probes and intermediate visual representations, as illustrated in Figure \ref{fig:method_diagram}. Unlike prior token-based approaches that reinterpret fixed encoder features ~\cite{blip2, flamingo, jia2022visualprompttuning}, \methodname~introduces a dedicated adaptation pathway that learns new visual representations from intermediate encoder layers while preserving the pretrained encoder parameters. In this section, we introduce how domain-adaptive VLMs are trained with \methodname.

% Our goal is to learn domain-specialized VLMs that avoid catastrophic forgetting without sacrificing the ability to learn domain-specific representations. To prevent catastrophic forgetting, we leave the vision encoder frozen during training and introduce \underline{\textbf{Vi}}sion \underline{\textbf{S}}pecialized \underline{\textbf{P}}robing (\textbf{\methodname}) to enable domain-specific visual representation learning. \methodname is a mechanism that augments the frozen vision encoder with a small set of learnable tokens, called \textit{visual probes}, and an interaction module that the probes with intermediate visual representations. In this section we introduce how we train domain-specialized VLMs with \methodname.

% Our goal is to train a domain-specialized VLM without updating the vision encoder to prevent catastrophic forgetting, while still enabling the model to learn domain-specific visual features. To achieve this, we introduce \underline{Vi}sion \underline{S}pecialized \underline{P}robing (\methodname), a mechanism that augments the frozen vision encoder with a small set of learnable tokens, called \textit{visual probes}, and an interaction module that enables these probes to interface with intermediate visual representations. In this section we introduce how we use \methodname to obtain domain-specialized VLMs.

% \noindent\textbf{Training Vision Language Models.}\quad

% \noindent\textbf{Training VLMs with \methodname.}\quad
\methodname\ augments the frozen vision encoder of a VLM with a compact set of $M$ learnable \emph{visual probes} $\mathbf{P}\in\mathbb{R}^{M\times d_v}$. The probes are trained to extract domain-specific spatio-temporal cues from intermediate representations of the vision encoder. To enable this extraction, a learnable \emph{interaction module} $\Phi^\ell$ inserted at each layer of the vision encoder conditions the probes on the hierarchical representations of the vision encoder at layer $\ell$:
\[
\mathbf{P}^{\ell+1} \;=\; \Phi^{\ell}(\mathbf{P}^{\ell},\mathbf{X}^{\ell}).
\]
Concretely, $\Phi^\ell$ is implemented as a vision-probe cross-attention between the visual embeddings and the probes at layer $\ell$. Let $(\boldsymbol{W_q}, \boldsymbol{W_k}, \boldsymbol{W_v})$ be the projection matrices in $\Phi^\ell$, then the probe update is
\[
\mathbf{P}^{\ell} \;=\; \mathrm{softmax}\!\left(\frac{\mathbf{P}^{\ell}\boldsymbol{W_q}^{\ell} (\mathbf{X}^{\ell}\boldsymbol{W_k}^{\ell})^\top}{\sqrt{d_v}}\right) (\mathbf{X}^{\ell} \boldsymbol{W_v}^{\ell}),
\]
Each $\Phi^\ell$ is parameterized independently, enabling layer-specific aggregation of low- to high-level visual semantics.
In contrast to the vision encoder of existing VLMs, which only learns spatial relationships through intra-frame self-attention, the visual probes attend to \emph{all} frames in the video, enabling them to learn complex spatio-temporal relationships specific to video understanding.
% While self-attention in the vision encoder operate independently over each frame in the visual sequence, the visual probes attend to \emph{all} spatio-temporal tokens; in some settings, such as robotic control, we restrict vision-probe cross-attention to spatial tokens only.
In some settings, such as robotic control, vision-probe cross-attentions are restricted to spatial tokens only.

After the final layer, the updated probes $\mathbf{P}^{L}$ are projected to the language embedding space via a dedicated connector $\mathcal{C}_{\text{probe}}$,
\[
\mathbf{Z}=\mathcal{C}_{\text{probe}}(\mathbf{P}^{L})\in\mathbb{R}^{M\times d_{\text{lm}}},
\]
and the VLM is trained with the standard autoregressive objective additionally conditioned on $\mathbf{Z}$:
\[
P(\mathbf{A}\mid \mathbf{E},\mathbf{Q},\mathbf{Z})
= \prod_{j=1}^{\mathrm{Len}}
P_{\boldsymbol{\theta}}\!\left(\mathbf{a}_j \mid \mathbf{E},\mathbf{Q},\mathbf{Z},\mathbf{A}_{<j}\right).
\]
Thus, the probes act as low-dimensional control knobs that bias learning toward domain-relevant structure and away from spurious artifacts. This is reinforced by applying updates through the probe connector, and through LoRA~\cite{lora} updates in the LLM embedding space, which confine parameter changes to a low-rank, probe-defined visual subspace that preserves generalizable behavior while enabling targeted specialization.

\section{Experiments}

We evaluate \methodname\ for effective domain adaptation and minimal forgetting. Section~\ref{sec:experiments_exp_settings} details the setup (architecture, training, metrics); Section~\ref{sec:experiments_targetsource_domains} reports results on egocentric, depth, and robotic-control targets; Section~\ref{sec:experiments_ablations} presents ablations and representation analyses of the probes and interaction modules.

\subsection{Experimental Setting}

\label{sec:experiments_exp_settings}
\noindent\textbf{VLM Architecture.}\quad
We consider a VLM architecture consisting of a SigLIP~\cite{zhai2023siglipv1} vision encoder, Qwen 2.5~\cite{qwen2025qwen25technicalreport} LLM, and a 2-layer MLP vision-language connector, with all modules initialized from the pretrained weights of VideoLLaMA3~\cite{damonlpsg2025videollama3}. The embedding dimension of the vision encoder is $d_v = 1152$, and the embedding dimension of the LLM is $d_{\text{lm}} = 3584$. We refer to this pretrained model as the \textit{base} VLM, and to models adapted to a target domain as \textit{expert} VLMs. To adapt the base VLM to a target domain, we perform finetuning on the target domain with a learning rate of $1\times10^{-5}$ for the LLM and vision-language connector, and a learning rate of $2\times10^{-6}$ for the vision encoder (when trainable). The model is finetuned on 4 NVIDIA H200 GPUs for 3 epochs when adapting to video domains, or 2 epochs when adapting to robotic control domains.

\noindent\textbf{\methodname~Details.}\quad
By default, \methodname~ operates at every layer of the vision encoder and employs $M=16$ visual probes unless otherwise stated. The visual probes are initialized from the normal distribution $\mathcal{N}(0,0.02)$. Each interaction module $\Phi^{\ell}$ is implemented as a multi-head cross-attention~\cite{vaswani2017attention}, and its weights are initialized from the self-attention weights of the vision encoder at layer $\ell$. During domain adaptation, we freeze the vision encoder and update only the visual probes, interaction modules, vision–language connectors, and the LLM’s LoRA parameters. For adaptation to video understanding domains, we update the LLM using LoRA ($r=16$), while the entire LLM is updated when adapting to the robotic control domain. We keep the vision--language connector trainable throughout, as is standard in VLM training~\cite{liu2023_llava, xue2025blip3}.

\noindent\textbf{Adaptation Metrics.}\quad
We evaluate the domain adaptation of VLMs across two dimensions: \textbf{(i)} their ``\textit{improvement}" on the target domain $\mathcal{T}$, and \textbf{(ii)} their ``\textit{retention}" on the source domain $\mathcal{S}$. Improvement on the target domain is measured as the performance difference between the expert and base VLMs on target domain benchmarks; retention is the corresponding difference on source domain benchmarks. If $\text{Acc}_\mathcal{D}$ denotes the average accuracy over all benchmarks within the domain $\mathcal{D}$, then the metrics are computed by:
\begin{align*}
\Delta_{\text{target}} &= \text{Acc}_{\text{target}}^{\text{expert}} - \text{Acc}_{\text{target}}^{\text{base}} \\
\Delta_{\text{source}} &= \text{Acc}_{\text{source}}^{\text{expert}} - \text{Acc}_{\text{source}}^{\text{base}}
\end{align*}

\subsection{Source and Target Domains}

\label{sec:experiments_mainresults}
\label{sec:experiments_targetsource_domains}
The source domain $\mathcal{S}$ is fixed throughout this paper: exocentric RGB videos of human actions reflecting the samples used to train generic VLMs for video representation learning. Our target domains $\mathcal{T}$ deliberately shift the input distribution, and consist of (1) egocentric video understanding, (2) depth-modality video understanding, and (3) robotic control. All data (videos and instructions) in our chosen target domain benchmarks were not used in the pretraining of the base VLM~\cite{damonlpsg2025videollama3}. We evaluate \methodname’s adaptation to each target while measuring retention of source domain competencies: (i) when adapting to egocentric video, exocentric understanding should be preserved; (ii) when adapting to depth video, RGB understanding should be preserved; and (iii) when adapting to robotic control, human-action understanding should be preserved.

%For each target domain, we select a source domain that reflects the training distribution of the base VLM, which is dominated by exocentric RGB videos of human actions, and that represents capabilities that should be preserved after adaptation.

\noindent\textbf{Training datasets.}\quad
For ego and depth video understanding domains, we adapt using EgoExo4D~\cite{cvpr2025egoexo4d}, a large-scale multi-view dataset containing time-synchronized egocentric and exocentric videos of skilled human activities. We utilize a total of 24,688 videos from the keystep recognition subset to generate 74,064 video instruction pairs. These instructions are recaptioned from the instruction pairs provided in \cite{reilly2025egoexo}. For the \textbf{egocentric} target domain, we adapt on 45{,}888 egocentric video-instruction pairs. For the \textbf{depth} target domain, we convert all exocentric RGB videos to depth using DepthAnythingV2~\cite{yang2024depthanythingv2} while keeping the language instructions unchanged, yielding 28{,}176 depth instruction pairs.

We perform adaptation to the \textbf{robotic control} domain in both simulated and real-world robot environments. In the \textit{simulated environment}, we leverage the training set of VIMA-Bench~\cite{jiang2023vimabench}. VIMA-Bench contains 17 object manipulation tasks with an action space comprising two 2D coordinates (for pick and place positions) and two quaternions (for rotation). Since the training set of VIMA-Bench lacks natural language instructions by default, we leverage the instruction pairs generated in LLaRA~\cite{xiang2025llara}, resulting in 13,922 instruction pairs across 7,995 action trajectories. In the \textit{real-world environment}, we collect a dataset using a 6-DoF xArm 7 robot arm deployed in a tabletop manipulation setting. This dataset, which we refer to as xArm-Det, contains 1,007 instruction pairs depicting novel objects and spatial configurations not present in simulation. During adaptation, we train jointly on VIMA-Bench and xArm-Det, resulting in a total of 14,929 instruction pairs. The large-scale simulated data enables the model to learn manipulation skills, while xArm-Det exposes the model to our novel robot environment. Illustrations of our real-world robot environment and examples from VIMA-Bench are provided in the Appendix.

\begin{table*}[th!]
\setlength{\tabcolsep}{5pt}
\renewcommand{\arraystretch}{1.2}
\centering
\caption{\textbf{Egocentric Video Understanding Experts.} Performance of adaptation strategies on the egocentric target domain and exocentric source domain.
\train~denotes trainable components, \frozen~denotes frozen components, and \train/\frozen~denotes LoRA updates.
$\Delta_{\text{target}}$ and $\Delta_{\text{source}}$ denote relative gains over the Base VLM.}
\resizebox{0.98\linewidth}{!}{
\begin{tabular}{cc|cccccc|ccccc|cc}
\hline
\multicolumn{2}{c|}{\cellcolor{gray!20}\textbf{Adaptation Strategy}} &
\multicolumn{6}{c|}{\cellcolor{gray!20}\textbf{Egocentric (Target)}} &
\multicolumn{5}{c|}{\cellcolor{gray!20}\textbf{Exocentric (Source)}} &
\multicolumn{2}{c}{\cellcolor{gray!20}\textbf{Metrics}} \\
\hline
% \textbf{VE} & \textbf{LLM} &
% \textbf{Act.} & \textbf{Task} & \textbf{HOI} & \textbf{Hand} & \textbf{EgoSch} & \textbf{Avg} &
% \textbf{NeXTQA} & \textbf{MME} & \textbf{MCQ} & \textbf{Desc} & \textbf{Avg} &
% $\Delta_\text{target}$ & $\Delta_\text{source}$ \\
\multirow{2}{*}{\textbf{VE}} & \multirow{2}{*}{\textbf{LLM}} &
\multirow{2}{*}{\shortstack{\textbf{Action}\\\textbf{Und.}}} & \multirow{2}{*}{\shortstack{\textbf{Task}\\\textbf{Regions}}} & \multirow{2}{*}{\textbf{HOI}} & \multirow{2}{*}{\shortstack{\textbf{Hand}\\\textbf{Ident.}}} & \multirow{2}{*}{\textbf{EgoSchema}} & \multirow{2}{*}{\textbf{Avg}} &
\multirow{2}{*}{\textbf{NeXTQA}} & \multirow{2}{*}{\shortstack{\textbf{Video}\\\textbf{MME}}} & \multirow{2}{*}{\shortstack{\textbf{ADL}\\\textbf{MCQ}}} & \multirow{2}{*}{\shortstack{\textbf{ADL}\\\textbf{Desc}}} & \multirow{2}{*}{\textbf{Avg}} &
\multirow{2}{*}{$\Delta_\text{target}$} & \multirow{2}{*}{$\Delta_\text{source}$} \\
& & & & & & & & & & & & & & \\
\hline

\rowcolor{gray!10}
\multicolumn{2}{c|}{\textbf{Base VLM}} &
75.37 & 74.88 & 75.56 & 65.38 & 60.98 & 70.43 &
\textbf{84.32} & \textbf{65.37} & 77.36 & 70.65 & 74.42 &
-- & -- \\
\hline

\multicolumn{15}{c}{\textbf{Finetuning Strategies (\train~Vision Language Connector)}} \\
\hline

\frozen & \frozen &
73.00 & 76.71 & 72.85 & 65.51 & 60.43 & 69.70 &
84.21 & 62.67 & 76.56 & 75.51 & 74.74 &
\textcolor{negativeorange}{-0.74} & \textcolor{codegreen}{+0.31} \\

\train & \frozen &
\underline{76.13} & \textbf{82.93} & 73.32 & 64.86 & 61.14 & \underline{71.68} &
83.87 & 61.41 & 77.05 & \underline{76.09} & 74.61 &
\textcolor{codegreen}{+1.24} & \textcolor{codegreen}{+0.18} \\

\train & \train &
73.28 & 82.68 & 72.96 & \textbf{65.77} & 60.31 & 71.00 &
82.34 & 64.26 & \underline{78.21} & 70.89 & 73.93 &
\textcolor{codegreen}{+0.57} & \textcolor{negativeorange}{-0.50} \\

\frozen & \frozen/\train &
73.49 & 74.27 & 74.50 & \underline{64.99} & \underline{61.52} & 69.75 &
84.24 & \underline{64.41} & 77.42 & 74.36 & \underline{75.11} &
\textcolor{negativeorange}{-0.68} & \underline{\textcolor{codegreen}{+0.68}} \\
\hline

\rowcolor{green!8}
\multicolumn{2}{c|}{\textbf{\methodname}} &
\textbf{81.28} & \underline{82.80} & \textbf{78.75} & 64.86 & \textbf{62.11} & \textbf{73.96} &
\underline{84.31} & 64.70 & \textbf{78.97} & \textbf{76.78} & \textbf{76.19} &
\textbf{\textcolor{codegreen}{+3.53}} & \textbf{\textcolor{codegreen}{+1.77}} \\

\hline
\end{tabular}}
\label{tab:ego2exo_sota}
\end{table*}

\subsubsection{Egocentric Video Understanding}
\label{sec:experiments_ego2exo_domain}
\noindent\textbf{Target and source benchmarks.}\quad
For evaluation on the \textbf{target domain}, we evaluate on the Ego-in-Exo Perception~\cite{reilly2025egoexo} and EgoSchema~\cite{mangalam2023egoschema} benchmarks. Ego-in-Exo Perception is derived from EgoExo4D and comprises 3,881 video question-answer (video-QA) pairs spanning four categories: action understanding (Action Und.), task-relevant region understanding (Task Regions), human-object interactions (HOI), and hand identification (Hand Ident.). Because it is derived from EgoExo4D, Ego-in-Exo Perception can be evaluated from either the egocentric or the exocentric viewpoint. For the ego target domain experiments, we report results using the egocentric videos, denoted as Ego-in-Exo Perception (Ego RGB). EgoSchema consists of 5,031 egocentric video-QA pairs derived from Ego4D~\cite{grauman2022ego4d}.

For evaluation on the \textbf{source domain}, we select benchmarks that measure exocentric video understanding capability. Specifically, we evaluate on the NeXTQA~\cite{xiao2021nextqa}, VideoMME~\cite{fu2025videomme}, and ADL-X~\cite{reilly2025llavidal} benchmarks. NeXTQA and VideoMME are general-purpose video-QA benchmarks built from web-scraped videos (e.g., from YouTube), with 8,564 QA pairs in NeXTQA and 2,700 QA pairs in VideoMME. ADL-X is a video-QA benchmark built from videos of activities of daily living, it contains a total of 10,561 multiple-choice questions (ADL-X MCQ) and 1,862 video description questions (ADL-X Desc) derived from various activities of daily living datasets~\cite{das2019toyotasmarthome, sigurdsson2016charades, jia2020lemma, dai2022toyotasmarthomeuntrimmed}.

\noindent\textbf{Results.}\quad
Table~\ref{tab:ego2exo_sota} reports results of adaptation to the egocentric viewpoint. Training only the vision-language connector or the connector together with LLM LoRA adapters does not lead to effective adaptation to the target domain ($\Delta_{\text{target}} < 1$). Updating all three modules (connector, vision encoder, and LLM) improves performance on the target domain by $\Delta_{\text{target}} = \textcolor{codegreen}{+0.57}$, but the large number of trainable parameters results in forgetting on the source benchmarks ($\Delta_{\text{source}} = \textcolor{negativeorange}{-0.50}$). In contrast, updating the connector and vision encoder alone slightly improves performance on the target domain and does not lead to forgetting on the source domain.
These results highlight that the core difficulty of domain adaptation in existing VLMs arises from the necessity of updating the vision encoder to learn domain-specific visual representations, which inevitably leads to forgetting of pretrained knowledge.
Our proposed \methodname~achieves the strongest adaptation performance, with the highest improvement on the target domain ($\Delta_{\text{target}}=\textcolor{codegreen}{+3.5}$) while simultaneously maintaining retention on the source benchmarks ($\Delta_{\text{source}}=\textcolor{codegreen}{+1.8}$). Interestingly, \methodname\ not only avoids catastrophic forgetting but also improves performance on some source benchmarks (e.g., ADL-X). We attribute this positive transfer to a multi-axis domain shift: although source and target differ in viewpoint (exocentric vs. egocentric), their action distributions overlap. ADL-X, while exocentric, encapsulates activities of daily living that closely aligns with the EgoExo4D action distribution, enabling beneficial cross-domain generalization.
\begin{table*}[th!]
\setlength{\tabcolsep}{5pt}
\renewcommand{\arraystretch}{1.2}
\centering
\caption{\textbf{Depth Video Understanding Experts.} Performance of adaptation strategies on the depth target domain and RGB source domain.
\train~denotes trainable components, \frozen~denotes frozen components, and \train/\frozen~denotes LoRA updates.
$\Delta_{\text{target}}$ and $\Delta_{\text{source}}$ denote relative gains over the Base VLM.}
\resizebox{0.98\linewidth}{!}{
\begin{tabular}{cc|ccccc|cccccc|cc}
\hline
\multicolumn{2}{c|}{\cellcolor{gray!20}\textbf{Adaptation Strategy}} &
\multicolumn{5}{c|}{\cellcolor{gray!20}\textbf{Depth (Target)}} &
\multicolumn{6}{c|}{\cellcolor{gray!20}\textbf{RGB (Source)}} &
\multicolumn{2}{c}{\cellcolor{gray!20}\textbf{Metrics}} \\
\hline
% \textbf{VE} & \textbf{LLM} &
% \textbf{Act.} & \textbf{Task} & \textbf{HOI} & \textbf{Hand} & \textbf{Avg} &
% \textbf{EgoRGB} & \textbf{NeXTQA} & \textbf{MME} & \textbf{MCQ} & \textbf{Desc} & \textbf{Avg} &
% $\Delta_\text{target}$ & $\Delta_\text{source}$ \\
\multirow{2}{*}{\textbf{VE}} & \multirow{2}{*}{\textbf{LLM}} &
\multirow{2}{*}{\shortstack{\textbf{Action}\\\textbf{Und.}}} & \multirow{2}{*}{\shortstack{\textbf{Task}\\\textbf{Regions}}} & \multirow{2}{*}{\textbf{HOI}} & \multirow{2}{*}{\shortstack{\textbf{Hand}\\\textbf{Ident.}}} & \multirow{2}{*}{\textbf{Avg}} & \multirow{2}{*}{\shortstack{\textbf{Ego-in-Exo}\\\textbf{(Exo RGB)}}} &
\multirow{2}{*}{\textbf{NeXTQA}} & \multirow{2}{*}{\shortstack{\textbf{Video}\\\textbf{MME}}} & \multirow{2}{*}{\shortstack{\textbf{ADL}\\\textbf{MCQ}}} & \multirow{2}{*}{\shortstack{\textbf{ADL}\\\textbf{Desc}}} & \multirow{2}{*}{\textbf{Avg}} &
\multirow{2}{*}{$\Delta_\text{target}$} & \multirow{2}{*}{$\Delta_\text{source}$} \\
& & & & & & & & & & & & & & \\
\hline

\rowcolor{gray!10}
\multicolumn{2}{c|}{\textbf{Base VLM}} &
34.73 & 50.61 & 35.06 & 63.06 & 45.86 &
66.27 & \textbf{84.32} & \textbf{65.37} & \textbf{77.36} & 70.65 & \underline{72.79} &
-- & -- \\
\hline

\multicolumn{15}{c}{\textbf{Finetuning Strategies (\train~Vision Language Connector)}} \\
\hline

\frozen & \frozen &
55.67 & 66.59 & \underline{62.46} & 64.49 & \underline{62.30} &
\underline{71.36} & 83.15 & 62.41 & 70.90 & 69.05 & 71.37 &
\underline{\textcolor{codegreen}{+16.44}} & \underline{\textcolor{negativeorange}{-1.42}} \\

\train & \frozen &
\textbf{57.20} & \underline{69.63} & 54.43 & \textbf{64.48} & 61.44 &
60.97 & 82.89 & 62.00 & 71.48 & 67.26 & 68.92 &
\textcolor{codegreen}{+15.57} & \textcolor{negativeorange}{-3.87} \\

\frozen & \train/\frozen &
42.94 & 53.54 & 43.92 & 63.96 & 51.09 &
60.97 & 83.73 & 64.19 & 72.19 & \underline{72.49} & 70.71 &
\textcolor{codegreen}{+5.23} & \textcolor{negativeorange}{-2.08} \\

\hline
\rowcolor{green!8}
\multicolumn{2}{c|}{\textbf{\methodname}} &
\underline{56.78} & \textbf{73.17} & \textbf{66.23} & \underline{64.35} & \textbf{65.13} &
\textbf{71.89} & \underline{83.91} & \underline{64.30} & \underline{76.59} & \textbf{76.47} & \textbf{74.63} &
\textbf{\textcolor{codegreen}{+19.27}} & \textbf{\textcolor{codegreen}{+1.84}} \\

\hline
\end{tabular}}
\label{tab:depth2rgb_sota}
\end{table*}

\subsubsection{Depth Video Understanding}

\noindent\textbf{Target and source benchmarks.}\quad
For evaluation on the \textbf{target domain}, we evaluate on the Ego-in-Exo Perception~\cite{reilly2025egoexo} benchmark. In the depth-adaptation setting, we train on depth maps of exocentric videos extracted with DepthAnythingV2~\cite{yang2024depthanythingv2} and evaluate on exocentric depth videos following~\cite{reilly2025egoexo}, denoted Ego-in-Exo Perception (Exo Depth). For the \textbf{source domain}, we use \textit{RGB} benchmarks of exocentric understanding: Ego-in-Exo Perception (Exo RGB), NeXTQA, VideoMME, and ADL-X.

\noindent\textbf{Results.}\quad
We present the results for adaptation to the depth modality in Table \ref{tab:depth2rgb_sota}. In contrast to the results on egocentric viewpoint adaptation, we find that all training strategies achieve improvements on the target domain, reflecting the disparity of the visual embedding space between the depth and RGB modalities. We find that this disparity leads to different behavior across training strategies. Jointly updating the vision encoder and the vision-language connector preserves source performance for egocentric adaptation but causes severe catastrophic forgetting under depth adaptation (\(\Delta_{\text{source}}=\textcolor{negativeorange}{-3.87}\)). This arises from the substantial encoder updates required to bridge RGB and depth, which overwrite RGB representations. In contrast, \methodname~preserves RGB features and source performance while achieving the largest target domain gains (\(\Delta_{\text{target}}=\textcolor{codegreen}{+19.27}\)).

\begin{table*}[th!]
\setlength{\tabcolsep}{5pt}
\renewcommand{\arraystretch}{1.2}
\centering
\caption{\textbf{Robot Control Experts (Simulation).} Performance of adaptation strategies on the robotic control target domain and human understanding source domain.
\train~denotes trainable components, \frozen~denotes frozen components.
$\Delta_{\text{target}}$ and $\Delta_{\text{source}}$ denote relative gains over the Base VLM.}
\resizebox{0.98\linewidth}{!}{
\begin{tabular}{cc|cccc|cccccc|cc}
\hline
\multicolumn{2}{c|}{\cellcolor{gray!20}\textbf{Adaptation Strategy}} &
\multicolumn{4}{c|}{\cellcolor{gray!20}\textbf{Robotic Control (Target)}} &
\multicolumn{6}{c|}{\cellcolor{gray!20}\textbf{Human Understanding (Source)}} &
\multicolumn{2}{c}{\cellcolor{gray!20}\textbf{Metrics}} \\
\hline
% \textbf{VE} & \textbf{LLM} &
% \textbf{L1} & \textbf{L2} & \textbf{L3} & \textbf{Avg} &
% \textbf{EgoRGB} & \textbf{NeXTQA} & \textbf{MME} & \textbf{MCQ} & \textbf{Desc} & \textbf{Avg} &
% $\Delta_\text{target}$ & $\Delta_\text{source}$ \\
\multirow{2}{*}{\textbf{VE}} & \multirow{2}{*}{\textbf{LLM}} &
\multirow{2}{*}{\textbf{L1}} & \multirow{2}{*}{\textbf{L2}} & \multirow{2}{*}{\textbf{L3}} &  \multirow{2}{*}{\textbf{Avg}} & \multirow{2}{*}{\shortstack{\textbf{Ego-in-Exo}\\\textbf{(Exo RGB)}}} &
\multirow{2}{*}{\textbf{NeXTQA}} & \multirow{2}{*}{\shortstack{\textbf{Video}\\\textbf{MME}}} & \multirow{2}{*}{\shortstack{\textbf{ADL}\\\textbf{MCQ}}} & \multirow{2}{*}{\shortstack{\textbf{ADL}\\\textbf{Desc}}} & \multirow{2}{*}{\textbf{Avg}} &
\multirow{2}{*}{$\Delta_\text{target}$} & \multirow{2}{*}{$\Delta_\text{source}$} \\
& & & & & & & & & & & & & \\
\hline

\rowcolor{gray!10}
\multicolumn{2}{c|}{\textbf{Base VLM}} &
0 & 0 & 0 & 0 &
\underline{66.27} & \textbf{84.32} & \textbf{65.37} & \textbf{77.36} & \textbf{70.65} & \textbf{72.79} &
-- & -- \\

\hline
\multicolumn{14}{c}{\textbf{Finetuning Strategies (\train~Vision Language Connector)}} \\
\hline
\train & \train &
\textbf{69.62} & 60.77 & 65.00 & \underline{65.13} &
56.92 & 83.24 & 62.74 & 52.21 & 64.50 & 63.92 &
\underline{\textcolor{codegreen}{+65.13}} & \textcolor{negativeorange}{-8.87} \\
\train & \frozen &
63.46 & 63.08 & 68.75 & 65.10 &
59.42 & 83.16 & \underline{64.41} & 52.92 & 64.86 & 64.95 &
\textcolor{codegreen}{+65.10} & \underline{\textcolor{negativeorange}{-7.84}} \\
\hline
\rowcolor{green!8}
\multicolumn{2}{c|}{\textbf{\methodname}} &
\underline{67.69} & \textbf{65.77} & \textbf{70.00} & \textbf{67.82} &
\textbf{71.19} & \underline{83.71} & 63.67 & \underline{55.89} & \underline{66.62} & \underline{68.22} &
\textbf{\textcolor{codegreen}{+67.82}} & \textbf{\textcolor{negativeorange}{-4.58}} \\
\hline

\end{tabular}}
\label{tab:robot2human_simulation}
\end{table*}

\subsubsection{Robot Control} \label{robotic-control}

\noindent\textbf{Target and source benchmarks.}\quad
For evaluation on the \textbf{target domain}, we consider both simulated and real-world robotic environments. In simulation, we use the evaluation set of VIMA-Bench~\cite{jiang2023vimabench}, which organizes tasks into three levels of difficulty: L1 (Object Placement), where all objects have been seen during training; L2 (Novel Combination), where objects seen during training appear in new pairings or contexts; and L3 (Novel Objects), where objects entirely unseen during training are introduced. Together, these levels measure generalization from familiar training conditions to progressively more challenging distributions. In the real-world setting, we evaluate on three tabletop manipulation tasks:  T1) Place the \texttt{\{object\}} on the plate, T2) Pick up and rotate \texttt{\{object\}} by \texttt{\{angle\}}; and T3) Move all \texttt{\{color\}} objects onto the plate. Examples of each task and a list of objects used is provided in the Appendix. On these robotic control benchmarks, the reported accuracy corresponds to the success rate across all robot manipulation tasks.
For \textbf{source domain} evaluation of VLMs trained on both real and simulated robotic environments, we use the human-activity video benchmarks Ego-in-Exo (Exo RGB), NeXTQA, VideoMME, and ADL-X.

\begin{table}[b]
\setlength{\tabcolsep}{8pt}
\renewcommand{\arraystretch}{1.2}
\centering
\vspace{-0.25cm}
\caption{\textbf{Robot Control Experts (Real-world).} Performance on the robotic control target domain and human understanding source domain.
\train~denotes trainable components and \frozen~denotes frozen components.}
\vspace{-0.5cm}
\resizebox{\linewidth}{!}{
\begin{tabular}{cc|cccc|cc}
\hline
\multicolumn{2}{c|}{\cellcolor{gray!20}\textbf{Adaptation Strategy}} &
\multicolumn{4}{c|}{\cellcolor{gray!20}\textbf{Robotic Control (Target)}} &
\multicolumn{2}{c}{\cellcolor{gray!20}\textbf{Metrics}} \\
\hline
\textbf{VE} & \textbf{LLM} &
\textbf{T1} & \textbf{T2} & \textbf{T3} & \textbf{Avg} &
$\Delta_\text{target}$ & $\Delta_\text{source}$ \\
\hline

\multicolumn{8}{c}{\cellcolor{gray!10}\textit{Training data: VIMA-Bench}} \\
\hline

% \train & \train &
% \textbf{45.00} & 60.00 & 15.00 & 40.00 &
% \underline{+40.00} & -8.87 \\
% \rowcolor{green!8}
% \multicolumn{2}{c|}{\textbf{\methodname}} &
% 40.00 & \textbf{70.00} & \textbf{20.00} & \textbf{43.33} &
% \textbf{+43.33} & \textbf{-4.58} \\
% \hline
% \multicolumn{8}{c}{\cellcolor{gray!10}\textit{Training data: VIMA-Bench + xArm-Det}} \\
% \hline
% \train & \train &
% 85.00 & 85.00 & 70.00 & 80.00 &
% \underline{+80.00} & -11.04 \\
% \rowcolor{green!8}
% \multicolumn{2}{c|}{\textbf{\methodname}} &
% \textbf{100.00} & \textbf{100.00} & \textbf{90.00} & \textbf{96.67} &
% \textbf{+96.67} & \textbf{-11.00} \\

\train & \train &
\textbf{45.00} & 60.00 & 15.00 & 40.00 &
\underline{\textcolor{codegreen}{+40.00}} & \textcolor{negativeorange}{-8.87} \\
\rowcolor{green!8}
\multicolumn{2}{c|}{\textbf{\methodname}} &
40.00 & \textbf{70.00} & \textbf{20.00} & \textbf{43.33} &
\textbf{\textcolor{codegreen}{+43.33}} & \textbf{\textcolor{negativeorange}{-4.58}} \\
\hline
\multicolumn{8}{c}{\cellcolor{gray!10}\textit{Training data: VIMA-Bench + xArm-Det}} \\
\hline
\train & \train &
85.00 & 85.00 & 70.00 & 80.00 &
\underline{\textcolor{codegreen}{+80.00}} & \textcolor{negativeorange}{-11.04} \\
\rowcolor{green!8}
\multicolumn{2}{c|}{\textbf{\methodname}} &
\textbf{100.00} & \textbf{100.00} & \textbf{90.00} & \textbf{96.67} &
\textbf{\textcolor{codegreen}{+96.67}} & \textbf{\textcolor{negativeorange}{-11.00}} \\
\hline
\end{tabular}}
\label{tab:robot2human_realworld}
\end{table}

\begin{figure*}[t]
  \centering
  % ===== Col 1: Table =====
\begin{minipage}[t]{0.38\linewidth}
\leavevmode
\captionsetup{type=table,skip=1pt}
\setlength{\tabcolsep}{2pt}
\renewcommand{\arraystretch}{1.2}
\centering
\caption{\textbf{Comparison with other adaptation strategies.} Legend: \emph{VP} (visual probing with no interaction modules), \emph{Last-4} (train only the last 4 vision encoder layers), \emph{QFormer Style} (interaction modules placed after the last VE layer).}
\resizebox{\linewidth}{!}{%
\begin{tabular}{l|c|c|cc}
\hline
\rowcolor{gray!20}
\textbf{Method} & \textbf{Target} & \textbf{Source} & \multicolumn{2}{c}{\textbf{Adaptation Metrics}} \\
\hline
 & \textbf{Avg} & \textbf{Avg} & $\boldsymbol{\Delta_\text{target}}$ & $\boldsymbol{\Delta_\text{source}}$ \\
\hline

Base VLM & 70.43 & 74.42 & -- & -- \\ \hline

\emph{VP} & 65.57 & 75.05 & \textcolor{negativeorange}{-4.86} & \textcolor{codegreen}{+0.62} \\
LoRA (Full) & 69.85 & 75.35 & \textcolor{negativeorange}{-0.59} & \textcolor{codegreen}{+0.92} \\
\emph{Last-4} & 70.46 & 72.62 & \textcolor{codegreen}{+0.02} & \textcolor{negativeorange}{-1.80} \\
\emph{VPT}~\cite{jia2022visualprompttuning} & 71.36 & 74.97 & \textcolor{codegreen}{+0.93} & \textcolor{codegreen}{+0.55} \\
\emph{QFormer Style} & 70.99 & 75.03 & \textcolor{codegreen}{+0.56} & \textcolor{codegreen}{+0.61} \\
Model Tailor~\cite{zhu2024modeltailor} & 70.27 & 75.29 & \textcolor{negativeorange}{-0.16} & \textcolor{codegreen}{+0.86} \\
\rowcolor{green!8}
\methodname & \textbf{73.96} & \textbf{76.19} & \textbf{\textcolor{codegreen}{+3.53}} & \textbf{\textcolor{codegreen}{+1.77}} \\

\hline
\end{tabular}%
}
\label{tab:ablation_alternatives}
\end{minipage}
  \hfill
    \begin{minipage}[t]{0.6\linewidth}
      \leavevmode
      \centering

      \begin{subfigure}[t]{0.49\linewidth}
        \centering
        \includegraphics[width=\linewidth]{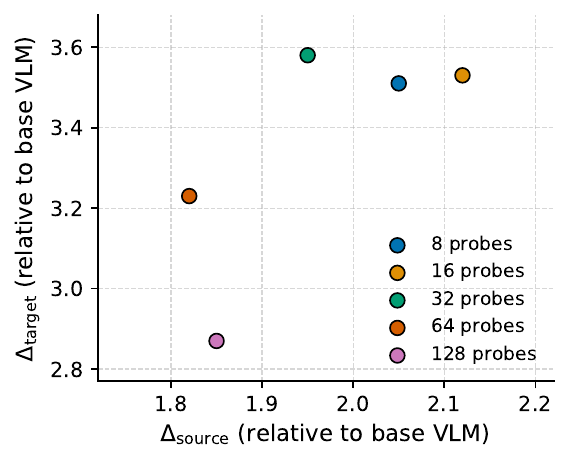}
        \caption{}
        \label{fig:ablation_numprobes}
      \end{subfigure}
      \hfill
      \begin{subfigure}[t]{0.49\linewidth}
        \centering
        \includegraphics[width=\linewidth]{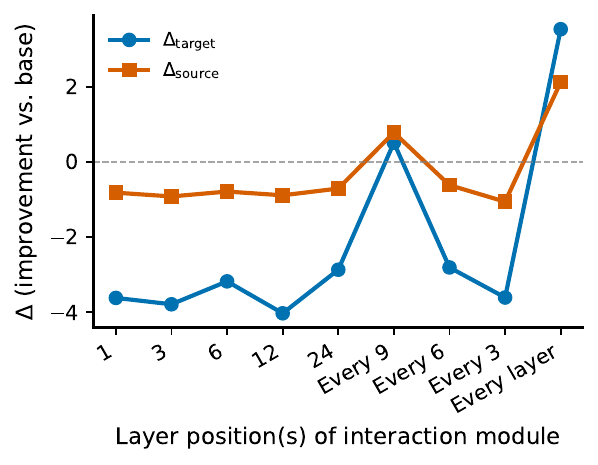}
        \caption{}
        \label{fig:ablation_layerpos}
      \end{subfigure}

      \captionof{figure}{\textbf{Ablation studies of \methodname~on the egocentric target domain.} (a) Effects of varying the number of visual probes within \methodname~(default=$16$). (b) Effect of interaction module placement position (default=\textit{Every layer}).}
      \label{fig:ablation_combined}
    \end{minipage}
    \vspace{-0.35cm}
\end{figure*}

\noindent\textbf{Results.}\quad
The results of adaptation to the robotic control domain are presented in Table \ref{tab:robot2human_simulation}.
The base VLM demonstrates weak performance on all robot control tasks, as its pretraining distribution lacks action trajectories (i.e., instruction data mapping from visual observations to robot actions). This lack of pretraining results in 0\% accuracy across all levels of VIMA-Bench, and is consistent with prior works~\cite{xiang2025llara}.
This highlights the extreme domain gap both in the visual space (robot observations vs. human videos) and in the language space (control actions vs. linguistic outputs) between the source and target domains. Similarly to the depth adaptation setting, we find that training the vision encoder improves performance on the target domain, but results in the worst source domain retention ($\Delta_{\text{source}}=\textcolor{negativeorange}{-8.87}$) of all robot control experts. In contrast, our proposed \methodname~achieves the best performance on the target domain ($\Delta_{\text{target}}=\textcolor{codegreen}{+67.82}$) while retaining the most source domain knowledge ($\Delta_{\text{source}}=\textcolor{negativeorange}{-4.58}$) compared to other experts, demonstrating the effectiveness of our method even when the gap between the source and target domains is very large. Also note that \methodname~operates on per-timestep images in these experiments; thus the visual probes consume the same visual tokens as the vision encoder, suggesting they extract domain-specific representations more effectively than the base encoder.

\noindent We further evaluate adaptation in the real-world setting using the xArm-Det dataset in Table~\ref{tab:robot2human_realworld}. We consider a \textit{transfer setting}, where the experts are trained only on VIMA-Bench and directly evaluated on xArm-Det, and the setting where the experts are jointly trained on both VIMA-Bench and xArm-Det. In both cases, our proposed \methodname~outperforms the vision encoder trained experts on target domain adaptation as well as source domain retention.

\subsection{Model Diagnosis and Analysis}

\label{sec:experiments_ablations}
% This section motivates \methodname's design through a diagnostic study of VLM adaptation to egocentric video understanding. We examine three factors: (i) the necessity of visual probes, (ii) the number of probes, and (iii) the placement of interaction modules within the vision encoder.
In this section, we motivate the design of \methodname~through a diagnostic study, and perform an analysis on the visual representations it learns. We investigate the number of visual probes, as well as the placement of interaction modules within the vision encoder. We then analyze the domain-specific representations learned by \methodname~through t-SNE and attention visualizations.

\noindent\textbf{Alternatives to learnable queries.}\quad Table~\ref{tab:ablation_alternatives} compares \methodname\ against alternative adaptation strategies. \textit{Visual Probes Only (VP)} trains only visual probes with their vision-language connector ($\mathcal{C}_{probe}$) without any interaction modules. \textit{Partial Encoder Training (Last-4)} makes the final four layers of the vision encoder trainable. \textit{QFormer-Style Compression} uses visual probes with interaction modules only at the vision encoder's final layer, mimicking Q-Former's compression approach~\cite{blip2}.
Notably, several of these alternatives underperform even the frozen base VLM on target tasks. We attribute this to their reliance on the pretrained visual pathway: they can recombine existing representations but struggle to introduce novel cues (e.g., egocentric or depth structure) that are absent from those representations. Full fine-tuning is more competitive on the target precisely because it can overwrite this pathway, at the cost of source-domain performance. \methodname\ occupies the middle ground, introducing a parallel probing pathway that decouples target-domain learning from the frozen pretrained pathway, and thereby learns new cues without overwriting the source representation.
\textit{Model Tailor}~\cite{zhu2024modeltailor} performs post-hoc domain adaptation by fusing parameter updates from a fine-tuned VLM back into the base VLM, modifying only the LLM parameters and leaving the vision encoder untouched. Training with QFormer-Style compression or only training with visual probes (\textit{VP}) underperforms compared to \methodname, indicating the importance of probe interactions at intermediate layers of the vision encoder to learn domain-specific features across multiple levels of abstraction. Similarly, training only the last four layers of the vision encoder, or training it with LoRA, also underperforms, highlighting that even partial parameter updates fail to capture domain-specific signals as effectively as \methodname's visual probes. Model Tailor also falls short in this setting, suggesting that approaches which do not leverage intermediate vision encoder representations struggle to learn domain-specific visual features.

\begin{table}[h]
    \setlength{\tabcolsep}{1pt}
    \renewcommand{\arraystretch}{1.2}
    \centering
    \vspace{-0.25cm}
    \captionof{table}{
    \textbf{Ablation on probe interaction scope.}
    Spatial-only probing restricts probe interactions within individual frames, while temporal-only probing restricts interactions across time. Full VisCoP jointly models spatial and temporal structure.}
    \resizebox{\linewidth}{!}{%
      \begin{tabular}{c|cc|cc}
        \hline
        \rowcolor{gray!20}
        \textbf{Interaction Module Scope} & \textbf{Target} & \textbf{Source} &
        \textbf{$\boldsymbol{\Delta_{\text{target}}}$ $\boldsymbol{(\uparrow)}$} &
        \textbf{$\boldsymbol{\Delta_{\text{source}}}$ $\boldsymbol{(\uparrow)}$} \\
        \hline
        Base VLM & 70.43 & 74.42 & -- & -- \\
        Spatial Only & 72.26 & 76.12 & \textcolor{codegreen}{+1.83} & \textcolor{codegreen}{+1.70} \\
        Temporal Only & 70.13 & 75.52 & \textcolor{negativeorange}{-0.30} & \textcolor{codegreen}{+1.10} \\
        Spatio-Temporal (\methodname) & \textbf{73.96} & \textbf{76.19} & \textcolor{codegreen}{\textbf{+3.53}} & \textcolor{codegreen}{\textbf{+1.77}} \\
        \hline
      \end{tabular}%
    }
    \label{tab:interaction_module_alternatives}
    % \vspace{-0.25cm}
\end{table}

\begin{figure*}[t]
  \centering
  % ===== Subfigure A =====
  \begin{subfigure}[t]{0.26\linewidth}
    \centering
    \includegraphics[width=\linewidth]{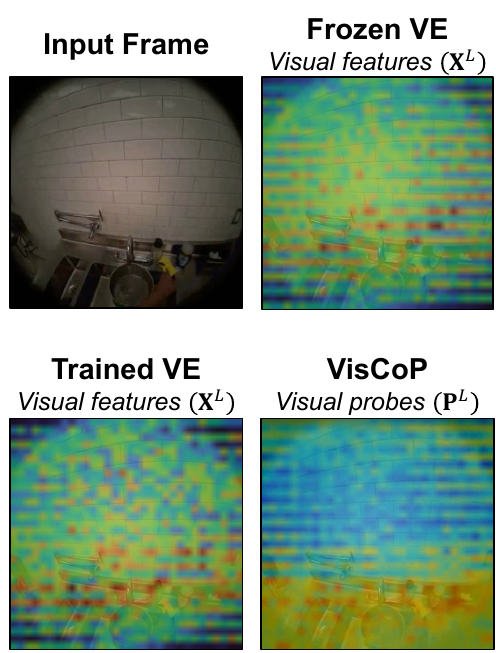}
    \caption{\textbf{Attention of visual features and probes.}}
    \label{fig:analysis_attn}
  \end{subfigure}
  \hfill
  % ===== Subfigure B =====
  \begin{subfigure}[t]{0.42\linewidth}
    \centering
    \includegraphics[width=\linewidth]{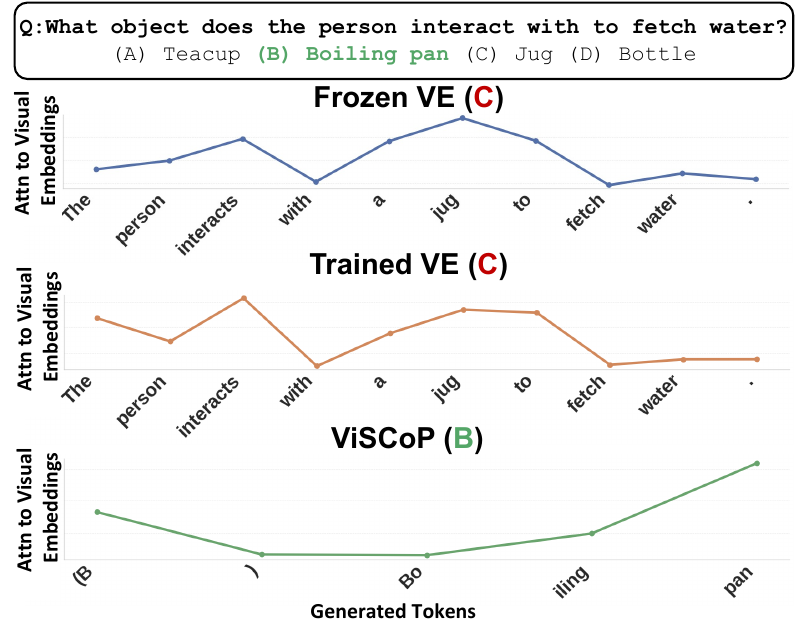}
    \caption{\textbf{Attentions of generated language tokens to visual embeddings.}}
    \label{fig:analysis_attnsums}
  \end{subfigure}
  \hfill
  % ===== Subfigure C =====
  \begin{subfigure}[t]{0.30\linewidth}
    \centering
    \includegraphics[width=\linewidth]{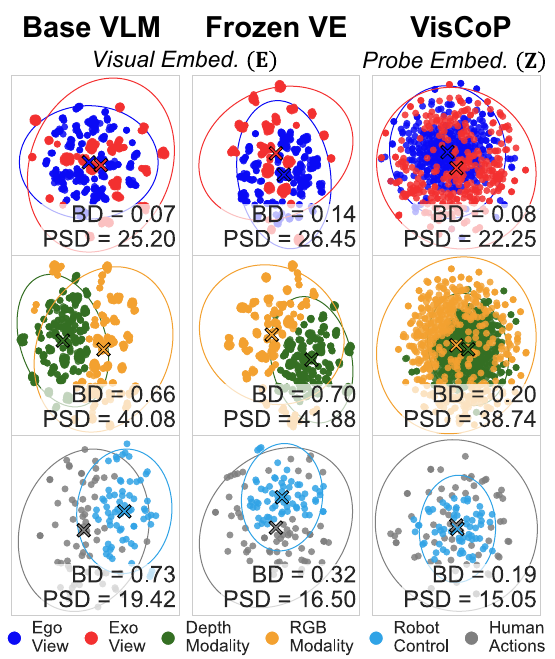}
    \caption{\textbf{t-SNE of source and target domain features.}}
    \label{fig:analysis_tsne}
  \end{subfigure}
  \caption{\textbf{Analysis of \methodname.} \textbf{(a)} Attentions between visual features and visual probes. \textbf{(b)} Attention of generated language tokens to visual embeddings. \textbf{(c)} t-SNE of visual and probe embeddings. Ellipses denote 95\% confidence regions of a fitted 2D Gaussian, and cross markers indicate the Gaussian means. Bhattacharyya distance (BD) and per-sample distance (PSD) are shown.}
  \label{fig:analysis_all}
  \vspace{-0.4cm}
\end{figure*}

\noindent\textbf{Effect of interaction module scope.}\quad
A key design choice in \methodname~is the scope of visual features that the probes interact through the interaction modules. In Table~\ref{tab:interaction_module_alternatives}, restricting probes to spatial-only or temporal-only interactions yields weaker adaptation performance, while full spatio-temporal interaction achieves substantially larger gains ($\Delta_{\text{target}}{=}\textcolor{codegreen}{+3.53}$). This result indicates that domain-relevant signals emerge from the joint spatio-temporal structure of video representations, and that enabling probes to integrate information across both dimensions is critical for effective video domain adaptation.

% Table~\ref{tab:ablation_alternatives} presents results for alternative approaches to incorporating visual signals during domain adaptation. Training only with visual probes (\textit{VP})
% underperforms compared to \methodname, indicating that visual probes alone are insufficient without interaction modules to integrate them effectively. A related baseline that augments probes with LoRA updates to both the vision encoder and LLM yields slightly stronger results but still lags behind \methodname, suggesting that lightweight parameter updates cannot close the domain gap effectively...

\iffalse
\begin{figure}[t]
\centering
\includegraphics[width=\linewidth]{content/analysis_domaintsne.pdf}
\caption{\textbf{t-SNE visualization of the three target domains.} The base VLM and the VLM adapted with a frozen vision encoder use pooled visual features, while \methodname{} uses pooled visual probes.}
\label{fig:analysis_domain_clustering}
\end{figure}
\fi

\begin{table}[h]
    \vspace{-0.25cm}
    \setlength{\tabcolsep}{6pt}
    \renewcommand{\arraystretch}{1.2}
    \centering
    \caption{\textbf{Computation overhead of VisCoP.}}
    \vspace{-0.25cm}
    \resizebox{\linewidth}{!}{
    \begin{tabular}{c|cccc}
         \hline
         \multirow{2}{*}{\textbf{Model}} & \multirow{2}{*}{\shortstack{\textbf{Max}\\\textbf{VRAM}}} & \multirow{2}{*}{\shortstack{\textbf{Latency}\\\textbf{(Full Model)}}} & \multirow{2}{*}{\shortstack{\textbf{Latency}\\\textbf{(Before LLM)}}} & \multirow{2}{*}{\shortstack{\textbf{Total Num}\\\textbf{Params.}}} \\
         & & & & \\
         \hline
         Base VLM & 24.4GB & 0.767s & 0.056s & 8.04B \\
         \methodname & 27.8GB & 0.417s & 0.069s & 8.21B \\
         \hline
    \end{tabular}}
    \label{tab:computation_overhead}
    \vspace{-0.25cm}
\end{table}

\noindent\textbf{Computation overhead of \methodname.}\quad
\methodname~ introduces only modest computational overhead relative to the base VLM, introducing only 2\% more parameters than the base VLM. In Table~\ref{tab:computation_overhead}, we compute the average VRAM usage and latency during inference on the Ego-in-Exo Perception benchmark, as well as the total parameter count. We find that \methodname~increases VRAM usage by 3.4GB and adds just 0.013s of inference latency in visual feature extraction (Before LLM). Interestingly, the inference latency of our model is lower than that of the base VLM. We attribute this to the base VLM producing longer, less focused responses, which increases total decoding time.

\noindent\textbf{Ablations on probes and interaction modules.}\quad
We study the effect of the number of visual probes and the placement of interaction modules (Figure \ref{fig:ablation_numprobes}, Figure \ref{fig:ablation_layerpos}). Probes consistently improve performance over the base VLM, with the best trade-off at 16 probes ($\Delta_{\text{target}}=\textcolor{codegreen}{+3.53}$, $\Delta_{\text{source}}=\textcolor{codegreen}{+2.12}$); larger probe counts offer no further gains and can reduce performance due to redundancy. For interaction modules, applying them at every encoder layer yields the strongest adaptation, while sparse placement (e.g., every 6 or 9 layers) provides weaker or inconsistent gains. These results highlight the importance of using a small number of probes with dense access to intermediate features.

\noindent\textbf{Visualizing attention in domain-adapted VLMs}\quad
In Figure~\ref{fig:analysis_attn}, we analyze attention maps of various VLM adaptation strategies to assess how different components capture domain-specific visual features. For both the frozen and trainable vision encoders, we visualize attention using attention rollout~\cite{abnar2020attentionrollout}, for \methodname~we visualize the attentions of the visual probes, averaged across all probes. The frozen vision encoder fails to focus consistently on relevant regions under the experimented domains, reflecting its limited ability to capture domain-specific features. The trained vision encoder yields sharper attention on the relevant regions, indicating its ability to learn domain-specific features, albeit at the cost of catastrophic forgetting of the source domain as shown in Section \ref{sec:experiments_mainresults}. In contrast, the visual probes of \methodname~have a sharp focus on the task-relevant regions, despite the vision encoder being frozen. This indicates that the probes alone are able to extract the domain-specific visual features necessary for adaptation. In Figure~\ref{fig:analysis_attnsums}, we visualize the attention of generated language tokens to visual embeddings. We find that \methodname~correctly responds to the query, with more focus given to tokens corresponding to relevant objects.

\noindent\textbf{Learning domain-specific representations.}\quad
Figure~\ref{fig:analysis_tsne} compares t-SNE embeddings of source and target domains across different VLMs. Circles represent individual samples, and ellipses denote 95\% confidence regions of fitted 2D Gaussians. For the egocentric and depth target domains, each source-target pair corresponds to time-synchronized videos of the same action. For the robot target domain, pairs correspond to pick-and-place actions performed by humans. Ideally, the embeddings of paired samples should lie closer together in the embedding space, reflecting alignment across the source and target domains. We quantify this using two metrics: the \emph{Bhattacharyya distance (BD)} computed between the Gaussians fitted to each domain, and the \emph{per-sample distance (PSD)}, defined as the mean Euclidean distance between paired embeddings across domains. We observe that the visual probes of \methodname\ learn stronger alignment between the source and target domains. Quantitatively, VisCoP yields the smallest Bhattacharyya distance and per-sample distance between paired source and target embeddings across all three target domains, indicating tighter cross-domain alignment than both the base VLM and the frozen-encoder baseline.
The effect is most pronounced under the largest shifts (depth and robot control), where the frozen encoder's paired embeddings remain widely separated while the probes draw them into closer correspondence.

\iffalse
\begin{figure*}[h!]
    \centering
    \includegraphics[width=\linewidth]{content/analysis_attn.pdf}
    \caption{\textbf{Attention of adaptation strategies.} \textbf{Left:} For frozen and trainable vision encoders, we visualize attention rollout; for \methodname, we visualize attentions of visual probes averaged across all probes. \textbf{Right:} Sum of attentions to visual tokens for each language token generated by the LLM.}
    \label{fig:analysis_ve-query-attention}
\end{figure*}
\fi

\begin{table}[h]
    \centering
    \caption{\textbf{Audio Understanding Experts.} Adaptation to a non-visual (spectrogram) target domain on MusicAVQA, with video understanding as the source domain.}
    \vspace{-0.25cm}
    \resizebox{\linewidth}{!}{
    \begin{tabular}{l|cccc}
         \hline
         % \multirow{2}{*}{\textbf{Method}} & \multirow{2}{*}{\shortstack{\textbf{Target}\\\textbf{Avg}}} & \multirow{2}{*}{\shortstack{\textbf{Source}\\\textbf{Avg}}} & \multirow{2}{*}{$\mathbf{\Delta}_\text{target}$} & \multirow{2}{*}{$\mathbf{\Delta}_\text{source}$} \\
         % & & & & \\
         \textbf{Method} & \textbf{Target Avg} & \textbf{Source Avg} & $\mathbf{\Delta}_\text{target}$ & $\mathbf{\Delta}_\text{source}$ \\
         \hline
         Base VLM & 28.3 & 57.5 & - & - \\
         \hline
         UE + LLM (LoRA) & 39.8 & 52.2 & \textcolor{codegreen}{+11.5} & \textcolor{negativeorange}{-5.3} \\
         \hline
         \textsc{VisCoP} & \textbf{40.2} & \textbf{56.8} & \textbf{\textcolor{codegreen}{+11.9}} & \textbf{\textcolor{negativeorange}{-0.7}} \\
         \hline
    \end{tabular}}
    \label{tab:audio_expert}
    \vspace{-0.25cm}
\end{table}

\noindent\textbf{Generalization beyond visual domains.} The target domains
considered thus far all shift the visual input distribution. To test whether
VisCoP generalizes to non-visual shifts, we adapt to \textit{audio
understanding}, where audio is represented as a spectrogram, using video
understanding as the source domain. Following OneLLM~\cite{han2024onellm}, we
repurpose the vision encoder into a universal encoder (UE) that processes both
audio and video, and instantiate VisCoP on top of it. We adapt and evaluate on
MusicAVQA~\cite{Li2022musicavqa}, treating its audio-centric questions as the
target domain and its video-centric questions as the source. As shown in
Table~\ref{tab:audio_expert}, VisCoP and the UE\,+\,LLM (LoRA) baseline are comparable
on the target domain ($\Delta_{\text{target}}=\textcolor{codegreen}{+11.9}$ vs.
$\textcolor{codegreen}{+11.5}$), but VisCoP nearly eliminates source-domain
forgetting ($\Delta_{\text{source}}=\textcolor{negativeorange}{-0.7}$ vs.
$\textcolor{negativeorange}{-5.3}$). This demonstrates VisCoP is not specific to visual shifts: it extracts domain-relevant cues while preventing forgetting even when the target modality is non-visual.

\vspace{-0.1cm}
\section{Conclusion}
\vspace{-0.1cm}

We introduced \methodname, a mechanism that extracts domain-specific visual features through probing of a frozen vision encoder to enable effective domain adaptation in VLMs and prevent catastrophic forgetting. VLMs equipped with \methodname\ achieve superior target domain performance, while maintaining strong source domain capabilities across cross-view, cross-modal, and cross-task adaptation scenarios. We will release all code, models, and evaluation protocols to facilitate future research.

\vspace{-0.1cm}
\section*{Acknowledgments}
\vspace{-0.1cm}

This work was supported in part by the National Science Foundation (IIS-2245652) and the University of North Carolina at Charlotte. Computational resources were provided by the NSF National AI Research Resource Pilot (NAIRR-240338) and the NCShare initiative. We would like to thank Hieu Le and other members of the Charlotte Vision Lab for their valuable insights and discussions.

{
    \small
    \bibliographystyle{ieeenat_fullname}
    \bibliography{main}

@inproceedings{mmbench,
      title={MMBench: Is Your Multi-modal Model an All-around Player?},
      author={Liu, Yuanzhan and Duan, Haodong and Zhang, Yuanhan and Li, Bo and Zhang, Songyang and Zhao, Wangbo and Yuan, Yike and Wang, Jiaqi and He, Conghui and Liu, Ziwei and Chen, Kai and Lin, Dahua},
      booktitle={European Conference on Computer Vision},
      year={2024}
}

@inproceedings{timesformer,
  author    = {Gedas Bertasius and Heng Wang and Lorenzo Torresani},
  title     = {Is Space-Time Attention All You Need for Video Understanding?},
  booktitle = {Proceedings of the International Conference on Machine Learning (ICML)},
  month     = {July},
  year      = {2021}
}

@article{liu2021videoswin,
  title   = {Video Swin Transformer},
  author  = {Ze Liu and Jia Ning and Yue Cao and Yixuan Wei and Zheng Zhang and Stephen Lin and Han Hu},
  journal = {2022 IEEE/CVF Conference on Computer Vision and Pattern Recognition (CVPR)},
  year    = {2021},
  pages   = {3192-3201}
}

@inproceedings{blip2,
  title={BLIP-2: Bootstrapping Language-Image Pre-training with Frozen Image Encoders and Large Language Models},
  author={Junnan Li and Dongxu Li and Silvio Savarese and Steven C. H. Hoi},
  booktitle={International Conference on Machine Learning},
  year={2023},
  url={https://api.semanticscholar.org/CorpusID:256390509}
}

@InProceedings{flamingo,
  author       = "Jean-Baptiste Alayrac and Jeff Donahue and Pauline Luc and Antoine Miech and Iain Barr and Yana Hasson and Karel Lenc and Arthur Mensch and Katherine Millican and Malcolm Reynolds and Roman Ring and Eliza Rutherford and Serkan Cabi and Tengda Han and Zhitao Gong and Sina Samangooei and Marianne Monteiro and Jacob L. Menick and Sebastian Borgeaud and Andy Brock and Aida Nematzadeh and Sahand Sharifzadeh and Mikołaj Bińkowski and Ricardo Barreira and Oriol Vinyals and Andrew Zisserman and Karén Simonyan",
  title        = "Flamingo: a Visual Language Model for Few-Shot Learning",
  booktitle    = "Advances in Neural Information Processing Systems",
  editor       = "S. Koyejo and S. Mohamed and A. Agarwal and D. Belgrave and K. Cho and A. Oh",
  pages        = "23716--23736",
  publisher    = "Curran Associates, Inc.",
  url          = "https://proceedings.neurips.cc/paper_files/paper/2022/file/960a172bc7fbf0177ccccbb411a7d800-Paper-Conference.pdf",
  volume       = "35",
  year         = "2022"
}

@inproceedings{liu2023_llava,
      title={Visual Instruction Tuning}, 
      author={Liu, Haotian and Li, Chunyuan and Wu, Qingyang and Lee, Yong Jae},
      booktitle={Advances in Neural Information Processing Systems (NeurIPS)},
      year={2023},
}

@inproceedings{
lora,
title={Lo{RA}: Low-Rank Adaptation of Large Language Models},
author={Edward J Hu and Yelong Shen and Phillip Wallis and Zeyuan Allen-Zhu and Yuanzhi Li and Shean Wang and Lu Wang and Weizhu Chen},
booktitle={International Conference on Learning Representations},
year={2022},
url={https://openreview.net/forum?id=nZeVKeeFYf9}
}

@inproceedings{vificlip,
  title     = {Finetuned CLIP models are efficient video learners},
  author    = {Rasheed, Hanoona and Khattak, Muhammad Uzair and Maaz, Muhammad and Khan, Salman and Khan, Fahad Shahbaz},
  booktitle = {The IEEE/CVF Conference on Computer Vision and Pattern Recognition},
  year      = {2023}
}

@inproceedings{yang2023aim,
  title={AIM: Adapting Image Models for Efficient Video Understanding},
  author={Yang, Taojiannan and Zhu, Yi and Xie, Yusheng and Zhang, Aston and Chen, Chen and Li, Mu},
  booktitle={International Conference on Learning Representations},
  year={2023},
  url={https://openreview.net/forum?id=CIoSZ_HKHS7}
}

@inproceedings{zang2023overcoming,
  title={Overcoming the Pitfalls of Vision-Language Model Finetuning for OOD Generalization},
  author={Zang, Yuhang and Goh, Hanlin and Susskind, Josh and Huang, Chen},
  booktitle={International Conference on Learning Representations},
  year={2024}
}

@article{lai2023lisa,
  title={LISA: Reasoning Segmentation via Large Language Model},
  author={Lai, Xin and Tian, Zhuotao and Chen, Yukang and Li, Yanwei and Yuan, Yuhui and Liu, Shu and Jia, Jiaya},
  journal={arXiv preprint arXiv:2308.00692},
  year={2023}
}

@article{Ranasinghe2024LearningTL,
  title={Learning to Localize Objects Improves Spatial Reasoning in Visual-LLMs},
  author={Kanchana Ranasinghe and Satya Narayan Shukla and Omid Poursaeed and Michael S. Ryoo and Tsung-Yu Lin},
  journal={2024 IEEE/CVF Conference on Computer Vision and Pattern Recognition (CVPR)},
  year={2024},
  pages={12977-12987},
  url={https://api.semanticscholar.org/CorpusID:269043025}
}

@article{minderer2023scaling,
  title={Scaling Open-Vocabulary Object Detection},
  author={Matthias Minderer, Alexey Gritsenko, Neil Houlsby},
  journal={NeurIPS},
  year={2023},
}

@inproceedings{li2024peftclip,
      title={Vision-Language Model Fine-Tuning via Simple Parameter-Efficient Modification},
      author={Li, Ming and Zhong, Jike and Li, Chenxin and Li, Liuzhuozheng and Lin, Nie and Sugiyama, Masashi},
      booktitle={Conference on Empirical Methods in Natural Language Processing},
      year={2024}
}

@article{damonlpsg2025videollama3,
  title={VideoLLaMA 3: Frontier Multimodal Foundation Models for Image and Video Understanding},
  author={Zhang, Boqiang and Li, Kehan and Cheng, Zesen and Hu, Zhiqiang and Yuan, Yuqian and Chen, Guanzheng and Leng, Sicong and Jiang, Yuming and Zhang, Hang and Li, Xin and Jin, Peng and Zhang, Wenqi and Wang, Fan and Bing, Lidong and Zhao, Deli},
  journal={arXiv preprint arXiv:2501.13106},
  year={2025},
  url = {https://arxiv.org/abs/2501.13106}
}

@article{Qwen2.5-VL,
  title={Qwen2.5-VL Technical Report},
  author={Bai, Shuai and Chen, Keqin and Liu, Xuejing and Wang, Jialin and Ge, Wenbin and Song, Sibo and Dang, Kai and Wang, Peng and Wang, Shijie and Tang, Jun and Zhong, Humen and Zhu, Yuanzhi and Yang, Mingkun and Li, Zhaohai and Wan, Jianqiang and Wang, Pengfei and Ding, Wei and Fu, Zheren and Xu, Yiheng and Ye, Jiabo and Zhang, Xi and Xie, Tianbao and Cheng, Zesen and Zhang, Hang and Yang, Zhibo and Xu, Haiyang and Lin, Junyang},
  journal={arXiv preprint arXiv:2502.13923},
  year={2025}
}

@misc{qwen2025qwen25technicalreport,
      title={Qwen2.5 Technical Report}, 
      author={{Qwen Team} and Yang, An and Yang, Baosong and Zhang, Beichen and Hui, Binyuan and Zheng, Bo and Yu, Bowen and Li, Chengyuan and Liu, Dayiheng and Huang, Fei and Wei, Haoran and Lin, Huan and Yang, Jian and Tu, Jianhong and Zhang, Jianwei and Yang, Jianxin and Yang, Jiaxi and Zhou, Jingren and Lin, Junyang and Dang, Kai and Lu, Keming and Bao, Keqin and Yang, Kexin and Yu, Le and Li, Mei and Xue, Mingfeng and Zhang, Pei and Zhu, Qin and Men, Rui and Lin, Runji and Li, Tianhao and Tang, Tianyi and Xia, Tingyu and Ren, Xingzhang and Ren, Xuancheng and Fan, Yang and Su, Yang and Zhang, Yichang and Wan, Yu and Liu, Yuqiong and Cui, Zeyu and Zhang, Zhenru and Qiu, Zihan},
      year={2025},
      eprint={2412.15115},
      archivePrefix={arXiv},
      primaryClass={cs.CL},
      url={https://arxiv.org/abs/2412.15115}, 
}

@inproceedings{zhai2023siglipv1,
      title={Sigmoid Loss for Language Image Pre-Training}, 
      author={Zhai, Xiaohua and Mustafa, Basil and Kolesnikov, Alexander and Beyer, Lucas},
      booktitle={Proceedings of the IEEE/CVF International Conference on Computer Vision},
      year={2023},
      pages={}, 
      organization={IEEE}
}

@inproceedings{vaswani2017attention,
      title={Attention Is All You Need}, 
      author={Vaswani, Ashish and Shazeer, Noam and Parmar, Niki and Uszkoreit, Jakob and Jones, Llion and Gomez, Aidan N. and Kaiser, Lukasz and Polosukhin, Illia},
      booktitle={Advances in Neural Information Processing Systems},
      year={2017}
}

@inproceedings{reilly2025egoexo,
  title     = {From My View to Yours: Learning Egocentric Cues from Exocentric Video using Privileged Egocentric Supervision},
  author    = {Dominick Reilly and Manish Kumar Govind and Le Xue and Srijan Das},
  booktitle = {Proceedings of the European Conference on Computer Vision (ECCV)},
  year      = {2026}
}

@inproceedings{cvpr2025egoexo4d,
      title={Ego-Exo4D: Understanding Skilled Human Activity from First- and Third-Person Perspectives}, 
      author={Grauman, Kristen and Westbury, Andrew and Torresani, Lorenzo and Kitani, Kris and Malik, Jitendra and Afouras, Triantafyllos and Ashutosh, Kumar and Baiyya, Vijay and Bansal, Siddhant and Boote, Bikram and Byrne, Eugene and Chavis, Zach and Chen, Joya and Cheng, Feng and Chu, Fu-Jen and Crane, Sean and Dasgupta, Avijit and Dong, Jing and Escobar, Maria and Forigua, Cristhian and Gebreselasie, Abrham and Haresh, Sanjay and Huang, Jing and Islam, Md Mohaiminul and Jain, Suyog and Khirodkar, Rawal and Kukreja, Devansh and Liang, Kevin J and Liu, Jia-Wei and Majumder, Sagnik and Mao, Yongsen and Martin, Miguel and Mavroudi, Effrosyni and Nagarajan, Tushar and Ragusa, Francesco and Ramakrishnan, Santhosh Kumar and Seminara, Luigi and Somayazulu, Arjun and Song, Yale and Su, Shan and Xue, Zihui and Zhang, Edward and Zhang, Jinxu and Castillo, Angela and Chen, Changan and Fu, Xinzhu and Furuta, Ryosuke and Gonzalez, Cristina and Gupta, Prince and Hu, Jiabo and Huang, Yifei and Huang, Yiming and Khoo, Weslie and Kumar, Anush and Kuo, Robert and Lakhavani, Sach and Liu, Miao and Luo, Mi and Luo, Zhengyi and Meredith, Brighid and Miller, Austin and Oguntola, Oluwatumininu and Pan, Xiaqing and Peng, Penny and Pramanick, Shraman and Ramazanova, Merey and Ryan, Fiona and Shan, Wei and Somasundaram, Kiran and Song, Chenan and Southerland, Audrey and Tateno, Masatoshi and Wang, Huiyu and Wang, Yuchen and Yagi, Takuma and Yan, Mingfei and Yang, Xitong and Yu, Zecheng and Zha, Shengxin Cindy and Zhao, Chen and Zhao, Ziwei and Zhu, Zhifan and Zhuo, Jeff and Arbelaez, Pablo and Bertasius, Gedas and Crandall, David and Damen, Dima and Engel, Jakob and Farinella, Giovanni Maria and Furnari, Antonino and Ghanem, Bernard and Hoffman, Judy and Jawahar, C. V. and Newcombe, Richard and Park, Hyun Soo and Rehg, James M. and Sato, Yoichi and Savva, Manolis and Shi, Jianbo and Shou, Mike Zheng and Wray, Michael},
      booktitle={Proceedings of the IEEE/CVF Conference on Computer Vision and Pattern Recognition},
      year={2025}
}

@inproceedings{jiang2023vimabench,
      title={VIMA: General Robot Manipulation with Multimodal Prompts},
      author={Jiang, Yunfan and Gupta, Agrim and Zhang, Zichen and Wang, Guanzhi and Dou, Yongqiang and Chen, Yanjun and Fei-Fei, Li and Anandkumar, Anima and Zhu, Yuke and Fan, Linxi},
      booktitle={International Conference on Machine Learning},
      year={2023}
}

@inproceedings{xiang2025llara,
      title={LLaRA: Supercharging Robot Learning Data for Vision-Language Policy},
      author={Li, Xiang and Mata, Cristina and Park, Jongwoo and Kahatapitiya, Kumara and Jang, Yoo Sung and Shang, Jinghuan and Ranasinghe, Kanchana and Burgert, Ryan and Cai, Mu and Lee, Yong Jae and Ryoo, Michael S.},
      booktitle={International Conference on Learning Representations},
      year={2025}
}

@inproceedings{yang2024depthanythingv2,
      title={Depth Anything V2},
      author={Yang, Lihe and Kang, Bingyi and Huang, Zilong and Zhao, Zhen and Xu, Xiaogang and Feng, Jiashi and Zhao, Hengshuang},
      booktitle={Advances in Neural Information Processing Systems},
      year={2024}
}

@inproceedings{mangalam2023egoschema,
  title        = {EgoSchema: A Diagnostic Benchmark for Very Long-form Video Language Understanding},
  author       = {Mangalam, Karttikeya and Akshulakov, Raiymbek and Malik, Jitendra},
  booktitle    = {Proceedings of the Thirty-seventh Conference on Neural Information Processing Systems, Datasets and Benchmarks Track},
  year         = {2023}
}

@inproceedings{grauman2022ego4d,
  title        = {Ego4D: Around the World in 3,000 Hours of Egocentric Video},
  author       = {Grauman, Kristen and Westbury, Andrew and Byrne, Eugene and Chavis, Zachary and Furnari, Antonino and Girdhar, Rohit and Hamburger, Jackson and Jiang, Hao and Liu, Miao and Liu, Xingyu and Martin, Miguel and Nagarajan, Tushar and Radosavovic, Ilija and Ramakrishnan, Santhosh Kumar and Ryan, Fiona and Sharma, Jayant and Wray, Michael and Xu, Mengmeng and Xu, Eric Zhongcong and Zhao, Chen and Bansal, Siddhant and Batra, Dhruv and Cartillier, Vincent and Crane, Sean and Do, Tien and Doulaty, Morrie and Erapalli, Akshay and Feichtenhofer, Christoph and Fragomeni, Adriano and Fu, Qichen and Gebreselasie, Abrham and Gonzalez, Cristina and Hillis, James and Huang, Xuhua and Huang, Yifei and Jia, Wenqi and Khoo, Weslie and Kolar, Jachym and Kottur, Satwik and Kumar, Anurag and Landini, Federico and Li, Chao and Li, Yanghao and Li, Zhenqiang and Mangalam, Karttikeya and Modhugu, Raghava and Munro, Jonathan and Murrell, Tullie and Nishiyasu, Takumi and Price, Will and Ruiz Puentes, Paola and Ramazanova, Merey and Sari, Leda and Somasundaram, Kiran and Southerland, Audrey and Sugano, Yusuke and Tao, Ruijie and Vo, Minh and Wang, Yuchen and Wu, Xindi and Yagi, Takuma and Zhao, Ziwei and Zhu, Yunyi and Arbelaez, Pablo and Crandall, David and Damen, Dima and Farinella, Giovanni Maria and Fuegen, Christian and Ghanem, Bernard and Ithapu, Vamsi Krishna and Jawahar, C. V. and Joo, Hanbyul and Kitani, Kris and Li, Haizhou and Newcombe, Richard and Oliva, Aude and Park, Hyun Soo and Rehg, James M. and Sato, Yoichi and Shi, Jianbo and Shou, Mike Zheng and Torralba, Antonio and Torresani, Lorenzo and Yan, Mingfei and Malik, Jitendra},
  booktitle    = {Proceedings of the IEEE/CVF Conference on Computer Vision and Pattern Recognition},
  year         = {2022},
  pages        = {18995--19012}
}

@inproceedings{xiao2021nextqa,
  title        = {NExT-QA: Next Phase of Question-Answering to Explaining Temporal Actions},
  author       = {Xiao, Junbin and Shang, Xindi and Yao, Angela and Chua, Tat-Seng},
  booktitle    = {Proceedings of the IEEE/CVF Conference on Computer Vision and Pattern Recognition},
  year         = {2021},
  pages        = {9777--9786}
}

@inproceedings{fu2025videomme,
  title        = {Video-MME: The First-Ever Comprehensive Evaluation Benchmark of Multi-modal Large Language Models in Video Analysis},
  author       = {Fu, Chaoyou and Dai, Yuhan and Luo, Yongdong and Li, Lei and Ren, Shuhuai and Zhang, Renrui and Wang, Zihan and Zhou, Chenyu and Shen, Yunhang and Zhang, Mengdan and Chen, Peixian and Li, Yanwei and Lin, Shaohui and Zhao, Sirui and Li, Ke and Xu, Tong and Zheng, Xiawu and Chen, Enhong and Shan, Caifeng and He, Ran and Sun, Xing},
  booktitle    = {Proceedings of the IEEE/CVF Conference on Computer Vision and Pattern Recognition},
  year         = {2025}
}

@inproceedings{reilly2025llavidal,
  title        = {LLAVIDAL: A Large Language-Vision Model for Daily Activities of Living},
  author       = {Reilly, Dominick and Chakraborty, Rajatsubhra and Sinha, Arkaprava and Govind, Manish Kumar and Wang, Pu and Bremond, Francois and Xue, Le and Das, Srijan},
  booktitle    = {Proceedings of the IEEE/CVF Conference on Computer Vision and Pattern Recognition},
  year         = {2025}
}

@inproceedings{das2019toyotasmarthome,
  title        = {Toyota Smarthome: Real-World Activities of Daily Living},
  author       = {Das, Srijan and Dai, Rui and Koperski, Michal and Minciullo, Luca and Garattoni, Lorenzo and Bremond, Francois and Francesca, Gianpiero},
  booktitle    = {Proceedings of the IEEE/CVF International Conference on Computer Vision},
  year         = {2019},
  pages        = {833--842}
}

@inproceedings{sigurdsson2016charades,
  title        = {Hollywood in Homes: Crowdsourcing Data Collection for Activity Understanding},
  author       = {Sigurdsson, Gunnar A. and Varol, G{\"u}l and Wang, Xiaolong and Farhadi, Ali and Laptev, Ivan and Gupta, Abhinav},
  booktitle    = {Proceedings of the European Conference on Computer Vision},
  year         = {2016},
  pages        = {510--526}
}

@inproceedings{jia2020lemma,
  title        = {LEMMA: A Multi-view Dataset for Learning Multi-agent Multi-task Activities},
  author       = {Jia, Baoxiong and Chen, Yixin and Huang, Siyuan and Zhu, Yixin and Zhu, Song-Chun},
  booktitle    = {Proceedings of the European Conference on Computer Vision},
  year         = {2020},
  pages        = {767--783}
}

@article{dai2022toyotasmarthomeuntrimmed,
  title        = {Toyota Smarthome Untrimmed: Real-World Untrimmed Videos for Activity Detection},
  author       = {Dai, Rui and Das, Srijan and Sharma, Saurav and Minciullo, Luca and Garattoni, Lorenzo and Bremond, Francois and Francesca, Gianpiero},
  journal      = {IEEE Transactions on Pattern Analysis and Machine Intelligence},
  year         = {2022},
  doi          = {10.1109/TPAMI.2022.3169976}
}

@misc{openai2025thinkwithimages,
  title={Thinking with Images},
  author={{OpenAI}},
  year={2025},
  month={April},
  url={https://openai.com/index/thinking-with-images/},
  note={Accessed: Jan 1 2026}
}

@misc{zhang2024llavavideo,
    title     = {Video Instruction Tuning With Synthetic Data}, 
    author    = {Zhang, Yuanhan and Wu, Jinming and Li, Wei and Li, Bo and Ma, Zejun and Liu, Ziwei and Li, Chunyuan},
    year      = {2024},
    eprint    = {2410.02713},
    archivePrefix = {arXiv},
    primaryClass  = {cs.CV},
    url       = {https://arxiv.org/abs/2410.02713}
}

@inproceedings{chen2024sharegpt4video,
      title={ShareGPT4Video: Improving Video Understanding and Generation with Better Captions},
      author={Chen, Lin and Wei, Xilin and Li, Jinsong and Dong, Xiaoyi and Zhang, Pan and Zang, Yuhang and Chen, Zehui and Duan, Haodong and Lin, Bin and Tang, Zhenyu and Yuan, Li and Qiao, Yu and Lin, Dahua and Zhao, Feng and Wang, Jiaqi},
      booktitle={Advances in Neural Information Processing Systems},
      year={2024}
}

@misc{maaz2024videogptplus,
      title={VideoGPT+: Integrating Image and Video Encoders for Enhanced Video Understanding},
      author={Maaz, Muhammad and Rasheed, Hanoona and Khan, Salman and Khan, Fahad},
      year={2024},
      eprint={2406.09418},
      archivePrefix={arXiv}
}

@misc{rawal2024cinepile,
      title={CinePile: A Long Video Question Answering Dataset and Benchmark},
      author={Rawal, Ruchit and Saifullah, Khalid and Basri, Ronen and Jacobs, David and Somepalli, Gowthami and Goldstein, Tom},
      year={2024},
      eprint={2405.08813},
      archivePrefix={arXiv}
}

@misc{xue2025blip3,
      title={xGen-MM (BLIP-3): A Family of Open Large Multimodal Models},
      author={Xue, Le and Shu, Manli and Awadalla, Anas and Wang, Jun and Yan, An and Purushwalkam, Senthil and Zhou, Honglu and Prabhu, Viraj and Dai, Yutong and Ryoo, Michael S and Kendre, Shrikant and Zhang, Jieyu and Tseng, Shaoyen and Lujan-Moreno, Gustavo A and Olson, Matthew L and Hinck, Musashi and Cobbley, David and Lal, Vasudev and Qin, Can and Zhang, Shu and Chen, Chia-Chih and Yu, Ning and Tan, Juntao and Awalgaonkar, Tulika Manoj and Heinecke, Shelby and Wang, Huan and Choi, Yejin and Schmidt, Ludwig and Chen, Zeyuan and Savarese, Silvio and Niebles, Juan Carlos and Xiong, Caiming and Xu, Ran},
      year={2025},
      eprint={2408.08872},
      archivePrefix={arXiv}
}

@misc{meta2024llama3herdmodels,
      title={The Llama 3 Herd of Models},
      author={Meta},
      year={2024},
      eprint={2407.21783},
      archivePrefix={arXiv}
}

@misc{radford2021openaiclip,
      title={Learning Transferable Visual Models From Natural Language Supervision},
      author={Radford, Alec and Kim, Jong Wook and Hallacy, Chris and Ramesh, Aditya and Goh, Gabriel and Agarwal, Sandhini and Sastry, Girish and Askell, Amanda and Mishkin, Pamela and Clark, Jack and Krueger, Gretchen and Sutskever, Ilya},
      year={2021},
      eprint={2103.00020},
      archivePrefix={arXiv}
}

@inproceedings{abnar2020attentionrollout,
      title={Quantifying Attention Flow in Transformers},
      author={Abnar, Samira and Zuidema, Willem},
      booktitle={Annual Meeting of the Association for Computational Linguistics},
      year={2020}
}

@inproceedings{yao2023kgcoop,
      title={Visual-Language Prompt Tuning with Knowledge-guided Context Optimization},
      author={Yao, Hantao and Zhang, Rui and Xu, Changsheng},
      booktitle={Proceedings of the IEEE/CVF Conference on Computer Vision and Pattern Recognition},
      year={2023}
}

@inproceedings{zhu2023prograd,
      title={Prompt-aligned Gradient for Prompt Tuning},
      author={Zhu, Beier and Niu, Yulei and Han, Yucheng and Wu, Yue and Zhang, Hanwang},
      booktitle={Proceedings of the IEEE/CVF International Conference on Computer Vision},
      year={2023}
}

@inproceedings{zhou2022coop,
      title={Conditional Prompt Learning for Vision-Language Models},
      author={Zhou, Kaiyang and Yang, Jingkang and Loy, Chen Change and Liu, Ziwei},
      booktitle={Proceedings of the IEEE/CVF Conference on Computer Vision and Pattern Recognition},
      year={2022}
}

@inproceedings{gao2023clipadapter,
      title={CLIP-Adapter: Better Vision-Language Models with Feature Adapters},
      author={Gao, Peng and Geng, Shijie and Zhang, Renrui and Ma, Teli and Fang, Rongyao and Zhang, Yongfeng and Li, Hongsheng and Qiao, Yu},
      booktitle={International Journal of Computer Vision},
      year={2023}
}

@inproceedings{cheng2025domainadaptationvlms,
      title={On Domain-Adaptive Post-Training for Multimodal Large Language Models},
      author={Cheng, Daixuan and Huang, Shaohan and Zhu, Ziyu and Zhang, Xintong and Zhao, Wayne Xin and Luan, Zhongzhi and Dai, Bo and Zhang, Zhenliang},
      booktitle={Conference on Empirical Methods in Natural Language Processing Findings},
      year={2025}
}

@misc{chen2024huatuogpt_medicaladaptvlm,
      title={HuatuoGPT-Vision, Towards Injecting Medical Visual Knowledge into Multimodal LLMs at Scale},
      author={Chen, Junying and Gui, Chi and Ouyang, Ruyi and Gao, Anningzhe and Chen, Shunian and Chen, Guiming Hardy and Wang, Xidong and Zhang, Ruifei and Cai, Zhenyang and Ji, Ke and Yu, Guangjun and Wan, Xiang and Wang, Benyou},
      year={2024},
      eprint={2406.19280},
      archivePrefix={arXiv}
}

@inproceedings{mohbat2024llavachef,
      title={LLaVA-Chef: A Multi-modal Generative Model for Food Recipes},
      author={Mohbat, Fnu and Zaki, Mohammed J.},
      booktitle={ACM International Conference on Information and Knowledge Management},
      year={2024}
}

@inproceedings{li2023llavamed,
      title={LLaVA-Med: Training a Large Language-and-Vision Assistant for Biomedicine in One Day},
      author={Li, Chunyuan and Wong, Cliff and Zhang, Sheng and Usuyama, Naoto and Liu, Haotian and Yang, Jianwei and Naumann, Tristan and Poon, Hoifung and Gao, Jianfeng},
      booktitle={Advances in Neural Information Processing Systems},
      year={2023}
}

@inproceedings{ha2024domainadaptqformer,
    title     = {Fusion of Domain-Adapted Vision and Language Models for Medical Visual Question Answering}, 
    author    = {Ha, Cuong Nhat and Asaadi, Shima and Karn, Sanjeev Kumar and Farri, Oladimeji and Heimann, Tobias and Runkler, Thomas},
    booktitle = {Proceedings of the Clinical Natural Language Processing Workshop at the 2024 Conference of the North American Chapter of the Association for Computational Linguistics (NAACL)},
    year      = {2024}
}

@misc{ryoo2025xgenmmvidblip3videoneed32,
      title={xGen-MM-Vid (BLIP-3-Video): You Only Need 32 Tokens to Represent a Video Even in VLMs},
      author={Ryoo, Michael S. and Zhou, Honglu and Kendre, Shrikant and Qin, Can and Xue, Le and Shu, Manli and Park, Jongwoo and Ranasinghe, Kanchana and Savarese, Silvio and Xu, Ran and Xiong, Caiming and Niebles, Juan Carlos},
      year={2025},
      eprint={2410.16267},
      archivePrefix={arXiv}
}

@inproceedings{zohar2025apollo,
      title={Apollo: An Exploration of Video Understanding in Large Multimodal Models},
      author={Zohar, Orr and Wang, Xiaohan and Dubois, Yann and Mehta, Nikhil and Xiao, Tong and Hansen-Estruch, Philippe and Yu, Licheng and Wang, Xiaofang and Juefei-Xu, Felix and Zhang, Ning and Yeung-Levy, Serena and Xia, Xide},
      booktitle={Proceedings of the IEEE/CVF Conference on Computer Vision and Pattern Recognition},
      year={2025}
}

@inproceedings{
socratic,
title={Socratic Models: Composing Zero-Shot Multimodal Reasoning with Language},
author={Andy Zeng and Maria Attarian and brian ichter and Krzysztof Marcin Choromanski and Adrian Wong and Stefan Welker and Federico Tombari and Aveek Purohit and Michael S Ryoo and Vikas Sindhwani and Johnny Lee and Vincent Vanhoucke and Pete Florence},
booktitle={The Eleventh International Conference on Learning Representations },
year={2023},
url={https://openreview.net/forum?id=G2Q2Mh3avow}
}

@inproceedings{zhu2024modeltailor,
      title={Model Tailor: Mitigating Catastrophic Forgetting in Multi-modal Large Language Models},
      author={Zhu, Didi and Sun, Zhongyi and Li, Zexi and Shen, Tao and Yan, Ke and Ding, Shouhong and Kuang, Kun and Wu, Chao},
      booktitle={International Conference on Machine Learning},
      year={2024}
}

@inproceedings{li2021prefixtuning,
  title={Prefix-Tuning: Optimizing Continuous Prompts for Generation},
  author={Li, Xiang Lisa and Liang, Percy},
  booktitle={ACL},
  year={2021}
}

@inproceedings{jia2022visualprompttuning,
  title={Visual Prompt Tuning},
  author={Jia, Menglin and Tang, Luming and Chen, Bor-Chun and Cardie, Claire and Belongie, Serge and Hariharan, Bharath and Lim, Ser-Nam},
  booktitle={ECCV},
  year={2022}
}

@InProceedings{han2024onellm,
  title={OneLLM: One Framework to Align All Modalities with Language},
  author={Han, Jiaming and Gong, Kaixiong and Zhang, Yiyuan and Wang, Jiaqi and Zhang, Kaipeng and Lin, Dahua and Qiao, Yu and Gao, Peng and Yue, Xiangyu},
  booktitle = {Proceedings of the IEEE/CVF Conference on Computer Vision and Pattern Recognition (CVPR)},
  year={2024}
}

@inproceedings{li2022musicavqa,
  title     = {Learning to Answer Questions in Dynamic Audio-Visual Scenarios},
  author    = {Guangyao Li and Yake Wei and Yapeng Tian and Chenliang Xu and Ji-Rong Wen and Di Hu},
  booktitle = {Proceedings of the IEEE/CVF Conference on Computer Vision and Pattern Recognition (CVPR)},
  year      = {2022}
}

@inproceedings{evci2022head2toe,
  title     = {Head2Toe: Utilizing Intermediate Representations for Better Transfer Learning},
  author    = {Utku Evci and Vincent Dumoulin and Hugo Larochelle and Michael C. Mozer},
  booktitle = {Proceedings of the 39th International Conference on Machine Learning (ICML)},
  year      = {2022}
}
}

\clearpage
\maketitlesupplementary

\appendix
\section{Appendix}
\subsection{Details of Simulated Robot Control Experiments}
\label{sec:appendix_details_simulated_robot}
\begin{figure*}[htbp]
    \centering
       
    \includegraphics[width=1\textwidth]{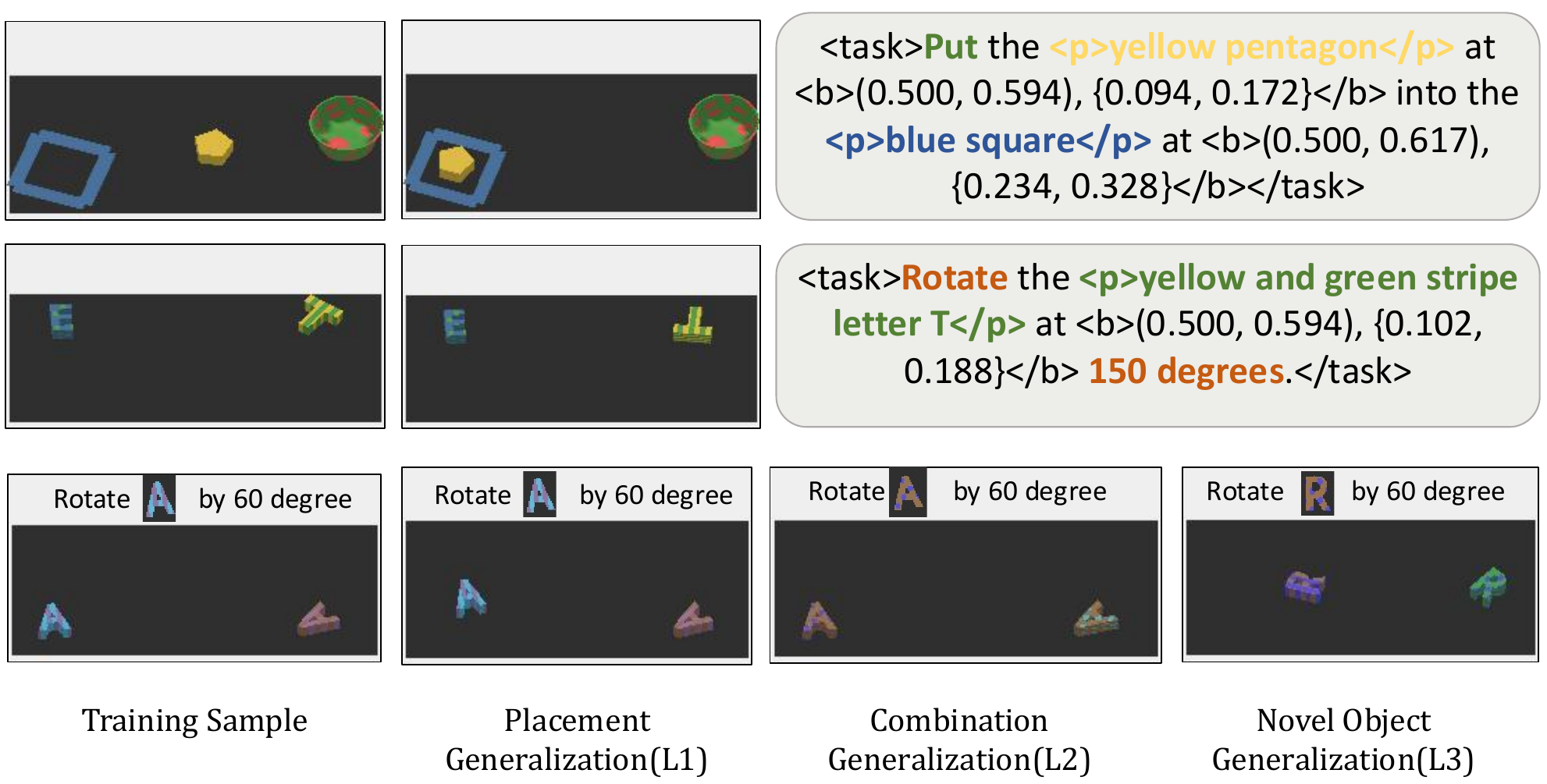}
    \caption{\textbf{Examples from VIMA and VIMA-Bench}. The first two rows show training examples, including the initial observations, final states, and task instructions. The bottom row illustrates the evaluation in VIMA-Bench, covering three levels of generalization.}
    \label{fig:VIMA-DinBC-Train-Eval-Example}

\end{figure*}

For our robot control simulation experiments, we use the VIMA-8K instruction set generated from the VIMA dataset, following \cite{xiang2025llara}. Figure \ref{fig:VIMA-DinBC-Train-Eval-Example} illustrates representative examples of training tasks - simple visual manipulation (top row) and rotation (middle row).  

For evaluation, we adopt the three levels of generalization defined in VIMA-Bench \cite{jiang2023vimabench}:  
- \textbf{L1 (Placement Generalization):} tasks where the object placements differ from those seen in the training set.  
- \textbf{L2 (Combination Generalization):} tasks requiring new combinations of objects not paired during training.  
- \textbf{L3 (Novel Object Generalization):} tasks involving completely unseen objects that were not present in the training data.

\begin{figure}[t]
    \centering
\includegraphics[width=0.25\textwidth,height=0.22\textwidth]{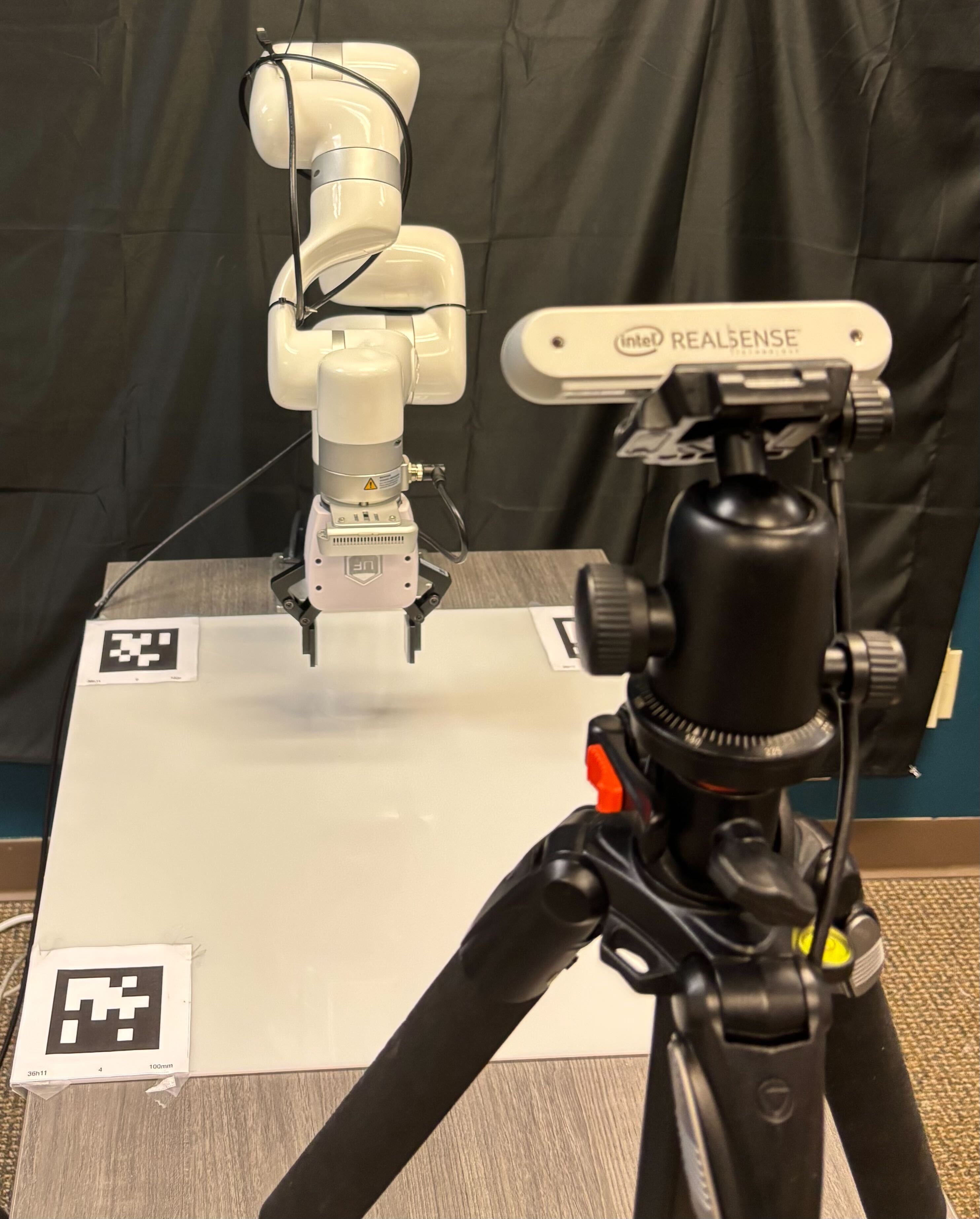}
    \caption{\textbf{Real robot setup.} Our setup uses an xArm7 robot arm and Intel RealSense D455 camera.}
    \label{fig:xarmsetup}
    
\end{figure}
\subsection{Details of Real-World Robotics Experiments}
\label{sec:appendix_details_realworld_robot}
We provide additional details of the experiments conducted in our novel robot environment, including the setup, data collection, and evaluation protocol.

\subsubsection{Real-Robot Setup}
Our setup consists of an xArm7 robotic arm with a gripper, tabletop, and an Intel RealSense D455 third person camera mounted in front of the arm to collect observations  as seen in Figure \ref{fig:xarmsetup}.
The action space of the end effector is  two 2D cartesian coordinates representing the pick and place poses, and two quaternions for rotations  similar to ~\cite{jiang2023vimabench}. We evaluated the effectiveness of our method mainly on three robot manipulative tasks:

\textbf{T1} : Place the \texttt{\{object\}} on the plate.
\textbf{T2} : Pickup and Rotate the \texttt{\{object\}} by \texttt{\{degree\}} degrees.
\textbf{T3} : Move all the  \texttt{\{colour\}} objects into the plate.

We uniformly sample \texttt{\{object\}} from a set of 10 toys : green apple, carrot, eggplant, banana, corn, grape, green pepper, tomato, strawberry, cucumber, clementine, and lemon. For T2, the target rotation angle is randomly selected from \{30\textdegree, 45\textdegree, 60\textdegree, 90\textdegree, 180\textdegree\}. For T3, the variable \texttt{{colour}} is chosen from four categories: \{red, orange, yellow, purple\}

\subsubsection{Real-Robot data collection}
% We can rephrase what is in LLaRA, probably dont need to show examples.

We collected 1,007 images with resolution 640 x 640 of a real-robot setup  with multiple objects scattered on the table. A one-shot object detection using  Owlv2 \cite{minderer2023scaling} is   applied to extract bounding boxes for each object. Based on these images and their corresponding bounding box annotations, we generate task instructions following the xArm-Det style similar to \cite{xiang2025llara}. 

\subsubsection{Evaluation protocol}\quad

\begin{figure*}[htbp]
    \centering
    \includegraphics[width=\textwidth]{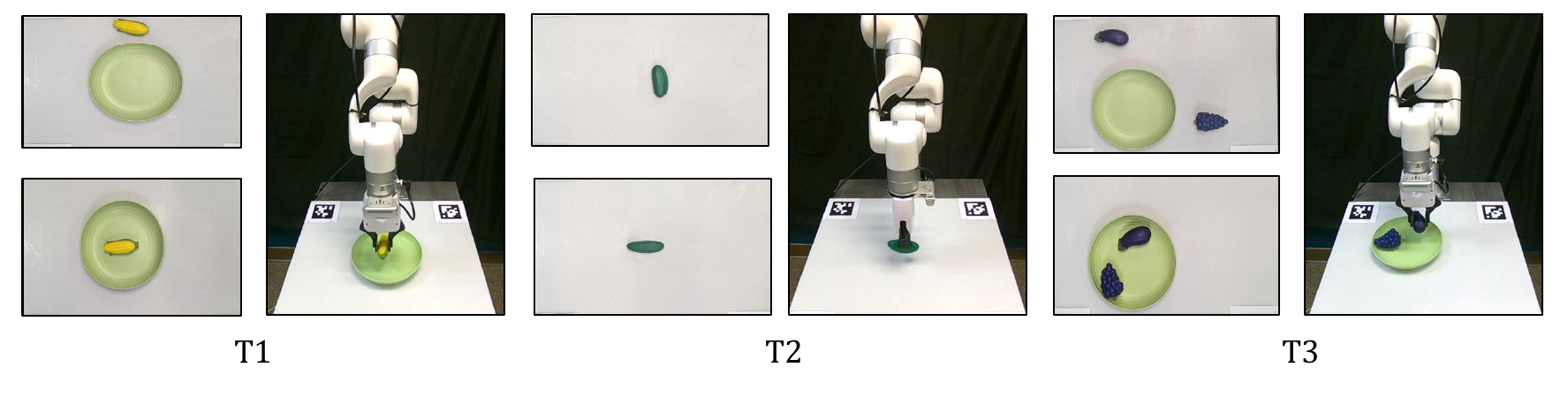}
    \caption{ \textbf{Visualization of the three real-world tasks.} Each column shows the initial state (top) and the corresponding final state (bottom), along with the robot execution (from left to right): \textbf{T1} (place the corn on the plate), \textbf{T2} (rotate the cucumber by $90^{\circ}$), and \textbf{T3} (move all purple objects into the plate).}
    \label{fig:real_task_vis}
\end{figure*}

All three tasks are evaluated under two settings: zero-shot and joint training. The observation space is illustrated in Figure \ref{fig:real_task_vis}. In zero-shot setting, we use the  models  trained on VIMA-8K where as  in the joint training setting, we finetune VLM jointly on both VIMA-8K and collected xArm-Det data. For each task, we conduct 20 trials with objects placed at random initial positions on the table. Each episode is limited to a maximum of 4 steps. We report the average success rate across all trials as performance metric  and  below are the success criteria for each task that we follow :

T1 : A trial is considered successful if at least 50\% of the object lies inside the plate.

T2 :  A trail is successful by  visually verifying whether the object has been rotated to the specified target angle.

T3 : A trial is successful only if all objects of the specified color are moved into the plate; otherwise, it is a failure.

\begin{figure*}[]
\centering
\begin{minipage}{0.245\textwidth}\centering
\includegraphics[width=\textwidth]{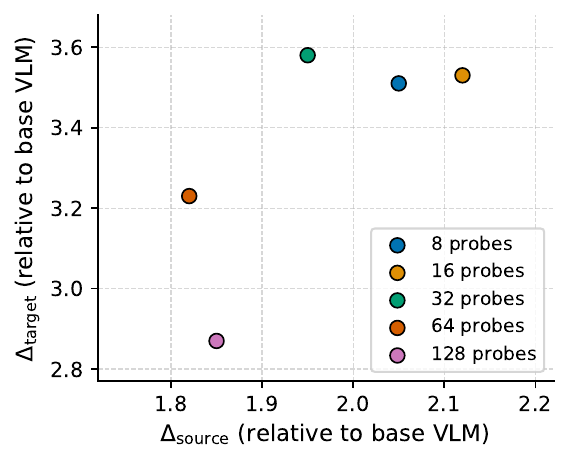}\\
(a)
\end{minipage}\hfill
\begin{minipage}{0.245\textwidth}\centering
\includegraphics[width=\textwidth]{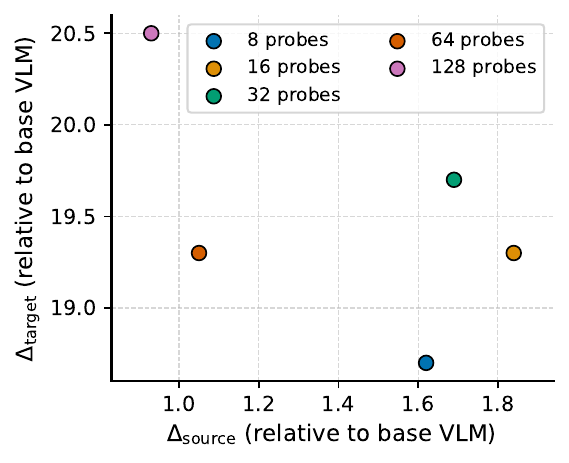}\\
(b)
\end{minipage}\hfill
\begin{minipage}{0.245\textwidth}\centering
\includegraphics[width=\textwidth]{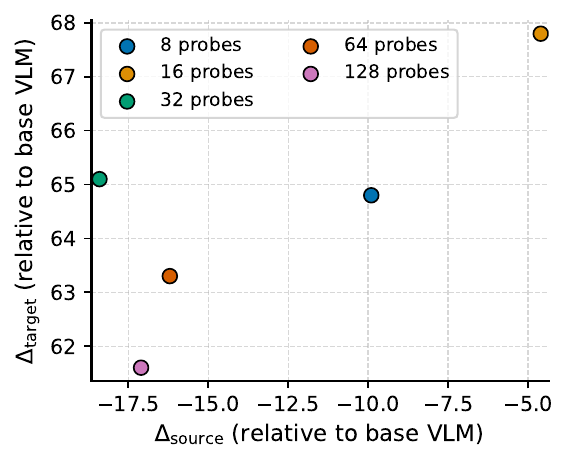}\\
(c)
\end{minipage}\hfill
\begin{minipage}{0.245\textwidth}\centering
\includegraphics[width=\textwidth]{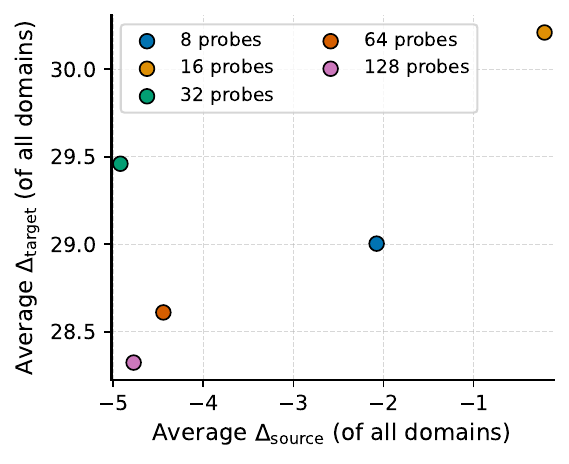}\\
(d)
\end{minipage}

\caption{\textbf{Effect of number of visual probes across domains.} (a) Egocentric viewpoint, (b) Depth modality, (c) Robotic control, (d) Average over all three target domains.}
\label{fig:analysis_probe_counts}

\end{figure*}

\subsection{Expanded Experimental Results}
\subsection{Is \methodname\ simply regularizing drift from the source model?} 
A natural hypothesis is that \methodname's source retention stems from limiting drift away from the base VLM, which dedicated retention objectives could achieve directly. We test this on the depth target domain with two retention-focused baselines: (i)~\textit{source-data replay}, adding 5\%, 10\%, and 25\% source-domain samples to the adaptation data, and (ii)~\textit{KL regularization}, adding a KL divergence between the output distributions of the adapted and base VLMs. As reported in Table~\ref{tab:retention}, replay improves source retention over target-only training but remains below the base VLM even at 25\% source data. KL regularization is a stronger baseline, improving both target accuracy and source retention. Yet VisCoP outperforms both on each axis, indicating that retention regularization alone does not account for its gains: the visual probes provide a more effective adaptation mechanism by exposing target-domain cues from intermediate representations without updating the pretrained encoder.

\begin{table}[t]
\centering
\caption{\textbf{Comparison to retention-focused baselines} on the depth target
domain. Replay augments adaptation with source-domain samples; KL-Reg. adds a KL
penalty against the base VLM. \methodname\ surpasses both on target and source.}
\label{tab:retention}
\begin{tabular}{lcccc}
\toprule
Method & Target Avg & Source Avg & $\Delta_{\text{target}}$ & $\Delta_{\text{source}}$ \\
\midrule
Base VLM     & 45.9 & 72.8 & -- & -- \\
Replay (0\%)  & 61.4 & 68.9 & \textcolor{codegreen}{+15.9} & \textcolor{negativeorange}{-3.9} \\
Replay (5\%)  & 60.8 & 69.9 & \textcolor{codegreen}{+14.9} & \textcolor{negativeorange}{-2.9} \\
Replay (10\%) & 60.8 & 70.0 & \textcolor{codegreen}{+14.9} & \textcolor{negativeorange}{-2.8} \\
Replay (25\%) & 61.6 & 70.0 & \textcolor{codegreen}{+15.7} & \textcolor{negativeorange}{-2.9} \\
KL-Reg.      & 63.9 & 73.0 & \textcolor{codegreen}{+17.9} & \textcolor{codegreen}{+0.2} \\
\methodname\       & \textbf{65.1} & \textbf{74.6} & \textcolor{codegreen}{\textbf{+19.3}} & \textcolor{codegreen}{\textbf{+1.8}} \\
\bottomrule
\end{tabular}
\end{table}

\subsubsection{Varying number of visual probes}
In Figure~\ref{fig:analysis_probe_counts}, we examine the effect of varying the number of visual probes across all three target domains. We verify that 16 probes provides a fair overall tradeoff across all target domains, retaining the most source-domain performance across all domains while remaining near the top in egocentric and depth adaptation, and performing substantially better than larger probe counts in the robotics setting. We attribute this to the fact that robotics tasks require the model to integrate visual cues with precise action semantics—when the number of probes becomes large, the additional probe signals tend to dominate the representation space, causing the LLM to overcommit to action-execution patterns and generate robotic command-like outputs even when inappropriate. In contrast, a smaller probe set provides focused domain-specific visual information without overwhelming the pretrained multimodal alignment, resulting in stronger performance and better retention.

\subsubsection{ADL-X Benchmark}
In this section, we present expanded results on the ADL-X benchmark across three target domains: \textbf{ego-video understanding} Table ~\ref{tab:egoexpert_adlx_expand}, \textbf{depth-video understanding} Table ~\ref{tab:depthexpert_adlx_expand}, and \textbf{robot control} Table \ref{tab:robotexpert_adlx_expand}. In addition, we provide comprehensive source-domain results for the real-world domain expert Table ~\ref{tab:robot2human_realworld_domain_expand}, as well as detailed ablation studies Table ~\ref{tab:ablations_extend}.

For the ADL-X description benchmark, we restrict evaluation to the  Charades Description ~\cite{reilly2025llavidal}.

\begin{table*}[t]
    \setlength{\tabcolsep}{5pt}
    \renewcommand{\arraystretch}{1.2}
    \centering
    \caption{\textbf{Performance of ego video expert on ADL-X Benchmark.}}
    \resizebox{0.98\linewidth}{!}{
    \begin{tabular}{ccc|ccccc|cccccc}
         \hline
         % Header row 1
         \multicolumn{3}{c|}{\cellcolor{gray!20}\textbf{Adaptation Strategy}} &
         \multicolumn{5}{c|}{\cellcolor{gray!20}\textbf{ADL-X MCQ}} &
         \multicolumn{6}{c}{\cellcolor{gray!20}\textbf{ADL-X Descriptions (Charades)}} \\
         \hline
         % Header row 2
         \textbf{VL-C} & \textbf{VE} & \textbf{LLM} &
         \textbf{Charades AR} & \textbf{Smarthome AR} &
         \textbf{TSU TC} & \textbf{LEMMA TC} & \textbf{Avg} &
         \textbf{Cor} & \textbf{Do} & \textbf{Ctu} & \textbf{Tu} & \textbf{Con} & \textbf{Avg} \\
         \hline
         % Data rows
         \multicolumn{3}{c|}{Base VLM} & 91.95 &  70.58 & 78.34 & 68.56 & 77.36  &73.50  & 73.74 & 75.78 & 68.59  & 61.61 & 70.64 \\
         \checkmark & \xmark & \xmark &   93.10 & 70.34 & 75.73 & 67.04 & 76.55 & 79.30 & 80.82 & 82.43 & 73.13 & 61.82 & 75.50 \\
         \checkmark & \checkmark & \xmark & 91.56 & 71.48 & 77.16 & 67.99 & 77.05 & 80.55 & 81.55 & 83.57 & 73.56 & 61.20 & 76.09 \\
         \checkmark & \xmark & LoRA & 92.39 & 71.50 & 77.59 & 68.18 & 77.41 & 78.54 & 77.34 & 81.70 & 73.45 & 60.74  & 74.36  \\
         \checkmark & \methodname & LoRA & 92.83 & 72.26 & 82.60 & 68.18 & 78.97 & 79.82 & 82.65 & 83.86 & 74.70 & 62.82 & 76.77   \\
         \hline
    \end{tabular}}
    \label{tab:egoexpert_adlx_expand}
\end{table*}

\begin{table*}[t]
    \setlength{\tabcolsep}{5pt}
    \renewcommand{\arraystretch}{1.2}
    \centering
    \caption{\textbf{Performance of depth video expert on ADL-X Benchmark.}}
    \resizebox{0.98\linewidth}{!}{
    \begin{tabular}{ccc|ccccc|cccccc}
         \hline
         % Header row 1
         \multicolumn{3}{c|}{\cellcolor{gray!20}\textbf{Adaptation Strategy}} &
         \multicolumn{5}{c|}{\cellcolor{gray!20}\textbf{ADL-X MCQ}} &
         \multicolumn{6}{c}{\cellcolor{gray!20}\textbf{ADL-X Descriptions (Charades)}} \\
         \hline
         % Header row 2
         \textbf{VL-C} & \textbf{VE} & \textbf{LLM} &
         \textbf{Charades AR} & \textbf{Smarthome AR} &
         \textbf{TSU TC} & \textbf{LEMMA TC} & \textbf{Avg} &
         \textbf{Cor} & \textbf{Do} & \textbf{Ctu} & \textbf{Tu} & \textbf{Con} & \textbf{Avg} \\
         \hline
         % Data rows
         \multicolumn{3}{c|}{Base VLM} & 91.95 &  70.58 & 78.34 & 68.56 & 77.36  &73.50  & 73.74 & 75.78 & 68.59  & 61.61 & 70.64 \\
         \checkmark & \xmark & \xmark & 90.84 & 56.26 & 71.51 & 64.96 & 70.89 & 71.22 & 75.31 & 75.95 & 65.80 & 56.96 & 69.05  \\
         \checkmark & \checkmark & \xmark & 90.90 & 54.87 & 73.65 & 66.47 & 71.47 & 69.96 & 73.60 & 73.83 & 64.04 & 54.82 & 67.25  \\
         \checkmark & \xmark & LoRA &  91.34 & 57.55 & 73.94 & 65.90 & 72.18 & 77.50 & 77.40 & 79.58 & 69.35 & 58.58 & 72.48 \\
         \checkmark & \methodname & LoRA & 93.60 & 63.79 & 81.71 & 67.23 & 76.58 & 78.51 & 84.68 & 84.07 & 74.67 & 60.41 & 76.47  \\
         \hline
    \end{tabular}}
    \label{tab:depthexpert_adlx_expand}
\end{table*}

\begin{table*}[t]
    \setlength{\tabcolsep}{5pt}
    \renewcommand{\arraystretch}{1.2}
    \centering
    \caption{\textbf{Performance of robot control expert  on ADL-X Benchmark.}}
    \resizebox{0.98\linewidth}{!}{
    \begin{tabular}{ccc|ccccc|cccccc}
         \hline
         % Header row 1
         \multicolumn{3}{c|}{\cellcolor{gray!20}\textbf{Adaptation Strategy}} &
         \multicolumn{5}{c|}{\cellcolor{gray!20}\textbf{ADL-X MCQ}} &
         \multicolumn{6}{c}{\cellcolor{gray!20}\textbf{ADL-X Descriptions (Charades)}} \\
         \hline
         % Header row 2
         \textbf{VL-C} & \textbf{VE} & \textbf{LLM} &
         \textbf{Charades AR} & \textbf{Smarthome AR} &
         \textbf{TSU TC} & \textbf{LEMMA TC} & \textbf{Avg} &
         \textbf{Cor} & \textbf{Do} & \textbf{Ctu} & \textbf{Tu} & \textbf{Con} & \textbf{Avg} \\
         \hline
         % Data rows
          \multicolumn{3}{c|}{Base VLM} &  91.95 &  70.58 & 78.34 & 68.56 & 77.36  &73.50  & 73.74 & 75.78 & 68.59  & 61.61 & 70.64 \\
         \checkmark & \checkmark & \checkmark & 78.05 & 36.24  & 36.39  & 58.14  & 52.21 & 66.16  & 68.30  & 70.66 & 61.78 & 55.6 & 64.50  \\
         \checkmark & \xmark & \checkmark & 78.88 & 39.81 &  35.61 & 57.38 & 52.95  & 66.54  & 68.95 & 71.03 & 62.58 & 55.15 & 64.85 \\
         \checkmark & \methodname & \checkmark & 90.96  & 45.77 & 38.76 & 48.1 & 55.89 &66.25 & 71.91 & 72.49 & 66.02 & 56.433  & 66.62  \\
         \hline
    \end{tabular}}
    \label{tab:robotexpert_adlx_expand}
\end{table*}

\begin{table*}
    \setlength{\tabcolsep}{8pt}
    \renewcommand{\arraystretch}{1.2}
    \centering
    \caption{\textbf{Expanded Robot Control Experts (Real-world)}}
    \resizebox{\linewidth}{!}{
    \begin{tabular}{ccc|cccc|cccccc|cc}
         \hline
         \multicolumn{3}{c|}{\cellcolor{gray!20}\textbf{Adaptation Strategy}} & \multicolumn{4}{c|}{\textbf{\cellcolor{gray!20}Robotic Control Benchmarks}} &
         \multicolumn{6}{c|}{\textbf{\cellcolor{gray!20}Human Understanding Benchmarks}} &
         \multicolumn{2}{c|}{\textbf{\cellcolor{gray!20}Adaptation Metrics}} \\
         \hline
         \multirow{2}{*}{\textbf{VL-C}} & \multirow{2}{*}{\textbf{VE}} & \multirow{2}{*}{\textbf{LLM}} & \multirow{2}{*}{\textbf{T1}} & \multirow{2}{*}{\textbf{T2}} & \multirow{2}{*} {\textbf{T3}} & \multirow{2}{*}{\textbf{Avg}} & 
         \multirow{2}{*}{\shortstack{\textbf{Ego-in-Exo}\\\textbf{(Exo RGB)}}} &
         \multirow{2}{*}{\textbf{NeXTQA}} &
         \multirow{2}{*}{\textbf{VideoMME}} &
         \multirow{2}{*}{\shortstack{\textbf{ADL-X}\\\textbf{MCQ}}} &
         \multirow{2}{*}{\shortstack{\textbf{ADL-X}\\\textbf{Desc}}} &
         \multirow{2}{*}{\textbf{Avg}} & 
         \textbf{$\boldsymbol{\Delta_{\text{target}}}$ } & \textbf{$\boldsymbol{\Delta_{\text{source}}}$ } \\
           & & &  &  &  &  & & & & & & & $\boldsymbol{(\uparrow)}$ &$\boldsymbol{(\uparrow)}$ \\
         \hline
         
         \hline
         \multicolumn{3}{c|}{Base VLM} & 0 & 0 & 0 & 0 & 66.27 & \textbf{84.32} & \textbf{65.37} & \textbf{77.36} & \textbf{70.65} & \textbf{72.79} & 
         - & - \\
         \multicolumn{15}{c}{\textit{\cellcolor{gray!10}Training data: VIMA-Bench}} \\
         \checkmark & \checkmark & \checkmark & \textbf{45.00} & 60.00 & 15.00 & 40.00 & 56.92 & 83.24 & 62.74 & 52.21 & 64.50 & 63.92 & \textcolor{codegreen}{+40.00} & \textcolor{negativeorange}{-8.87} \\
         \checkmark & \methodname & \checkmark & 40.00 & \textbf{70.00} & \textbf{20.00} & \textbf{43.33} & \textbf{71.19} & \underline{83.71} & \underline{63.67} & \underline{55.89} & 66.62 & \underline{68.22} & \textcolor{codegreen}{+\textbf{43.33}} & \textcolor{negativeorange}{-\textbf{4.58}} \\
         \hline
         \multicolumn{15}{c}{\textit{\cellcolor{gray!10}Training data: VIMA-Bench + xArm-Det}} \\
         \checkmark & \checkmark & \checkmark & 85.00 & 85.00 & 70.00 & 80.00 &  64.50	& 83.00 &	63.00	 & 36.04 &	62.24 &	61.76 & \textcolor{codegreen}{+80.00} & \textcolor{negativeorange}{-11.04} \\
         \checkmark & \methodname & \checkmark & \textbf{100.00} & \textbf{100.00} & \textbf{90.00} & \textbf{96.67} & 59.59 &	82.98 &	63.26 &	36.32 &	\underline{66.83} &	61.79 &  \textcolor{codegreen}{\textbf{+96.67}} & \textcolor{negativeorange}{\textbf{-11.00}} \\
         \hline
    \end{tabular}}
    \label{tab:robot2human_realworld_domain_expand}
\end{table*}

\begin{table*}[t]
    \setlength{\tabcolsep}{5pt} % default is 6pt
    \renewcommand{\arraystretch}{1.2}

    \centering
    \caption{\textbf{Comprehensive target-source domain results from the ablation study of \methodname}}
    \resizebox{0.98\linewidth}{!}{
    \begin{tabular}{ccc|cccccc|ccccc|cc}
         \hline
         % Row 1 (header level 1)
         \multicolumn{3}{c|}{\cellcolor{gray!20}\multirow{1}{*}{\textbf{Adaptation Strategy}}} &
         \multicolumn{6}{c|}{\cellcolor{gray!20}\textbf{Egocentric Benchmarks}} &
         \multicolumn{5}{c|}{\cellcolor{gray!20}\textbf{Exocentric Benchmarks}} &
         \multicolumn{2}{c|}{\cellcolor{gray!20}\textbf{Adaptation Metrics}} \\
         \cline{1-16}
         % Row 2 (header level 2)
         \multirow{3}{*}{\textbf{VL-C}} & \multirow{3}{*}{\textbf{VE}} & \multirow{3}{*}{\textbf{LLM}} &
         \multicolumn{4}{c}{\textit{Ego-in-Exo PerceptionMCQ (Ego RGB)}} & \multirow{3}{*}{\textbf{EgoSchema}} & \multirow{3}{*}{\textbf{Avg}} &
         \multirow{3}{*}{\textbf{NeXTQA}} &
         \multirow{3}{*}{\textbf{VideoMME}} & \multirow{3}{*}{\shortstack{\textbf{ADL-X}\\\textbf{MCQ}}}
         & \multirow{3}{*}{\shortstack{\textbf{ADL-X}\\\textbf{Desc}}} &
         \multirow{3}{*}{\textbf{Avg}} & \multirow{3}{*}{\shortstack{\textbf{$\boldsymbol{\Delta_{\text{target}}}$}\\\textbf{$\boldsymbol{(\uparrow)}$}}} &
         \multirow{3}{*}{\shortstack{\textbf{$\boldsymbol{\Delta_{\text{source}}}$}\\\textbf{$\boldsymbol{(\uparrow)}$}}} \\
         % Row 3 (new row - first part of split headers)
         & & &
         \multirow{2}{*}{\shortstack{\textbf{Action}\\\textbf{Und.}}} & \multirow{2}{*}{\shortstack{\textbf{Task}\\\textbf{Regions}}} & \multirow{2}{*}{\textbf{HOI}} & \multirow{2}{*}{\shortstack{\textbf{Hand}\\\textbf{Ident.}}} & & &
           &  &  &  &  &  &  \\
         % Row 4 (header level 3 - second part of split headers)
         & & &
          & & & & & &
           &  &  &  &  &  &  \\
         \hline
         \multicolumn{3}{c|}{Base VLM} & 75.37 & 74.88 & \underline{75.56} & \textbf{65.38} & 60.98 & 70.43 & \textbf{84.32} & \textbf{65.37} & 77.36 & 70.65 & \underline{74.42} & - & - \\
             \checkmark & VP & LoRA & 66.88 &	75.98 &	59.62&	63.84 &	61.54 &	65.57 &	84.22 &	64.37 &	77.86 &	73.73 &	75.05 &\textcolor{negativeorange}{-4.86}	& \textcolor{codegreen}{0.62} \\
         \checkmark & LoRA & LoRA & 73.76 &	75.24 &	73.55 &	\underline{64.99} &	61.68 &	69.85 &	84.22 &	64.48 &	77.52 &	75.17 &	75.35 & \textcolor{negativeorange}{-0.59} & \textcolor{codegreen}{0.92} \\
         \checkmark & last-4 & LoRA & 
73.35 &	\underline{77.93} &	73.32 &	65.25 &	\textbf{62.43} & 70.46 &	84.00 &	63.78 &	77.74 &	\underline{76.34} &	72.62 &	\textcolor{codegreen}{0.02} &	\textcolor{negativeorange}{-1.80} \\
         \checkmark & QFormer-Style & LoRA & \underline{75.99}	& 77.56 &	74.50 &	\textbf{65.38} &	\underline{61.54} &	\underline{70.99} &	84.13 &	64.44 &	\underline{78.43} &	73.13 &	75.03 &	\textcolor{codegreen}{0.56}	& \textcolor{codegreen}{0.61}  \\
         \checkmark & \methodname & LoRA & \textbf{81.28} & \textbf{82.80} & \textbf{78.75} & 64.86 & 62.11 & \textbf{73.96} & \underline{84.31} & \underline{64.70} & \textbf{78.97} & \textbf{76.78} & \textbf{76.19} & \textcolor{codegreen}{\textbf{+3.53}} & \textcolor{codegreen}{\textbf{+1.77}} \\
         \hline
    \end{tabular}}
    \label{tab:ablations_extend}
\end{table*}

\subsection{Qualitative results}
\begin{figure*}[t]
    \centering
    \includegraphics[width=\textwidth]{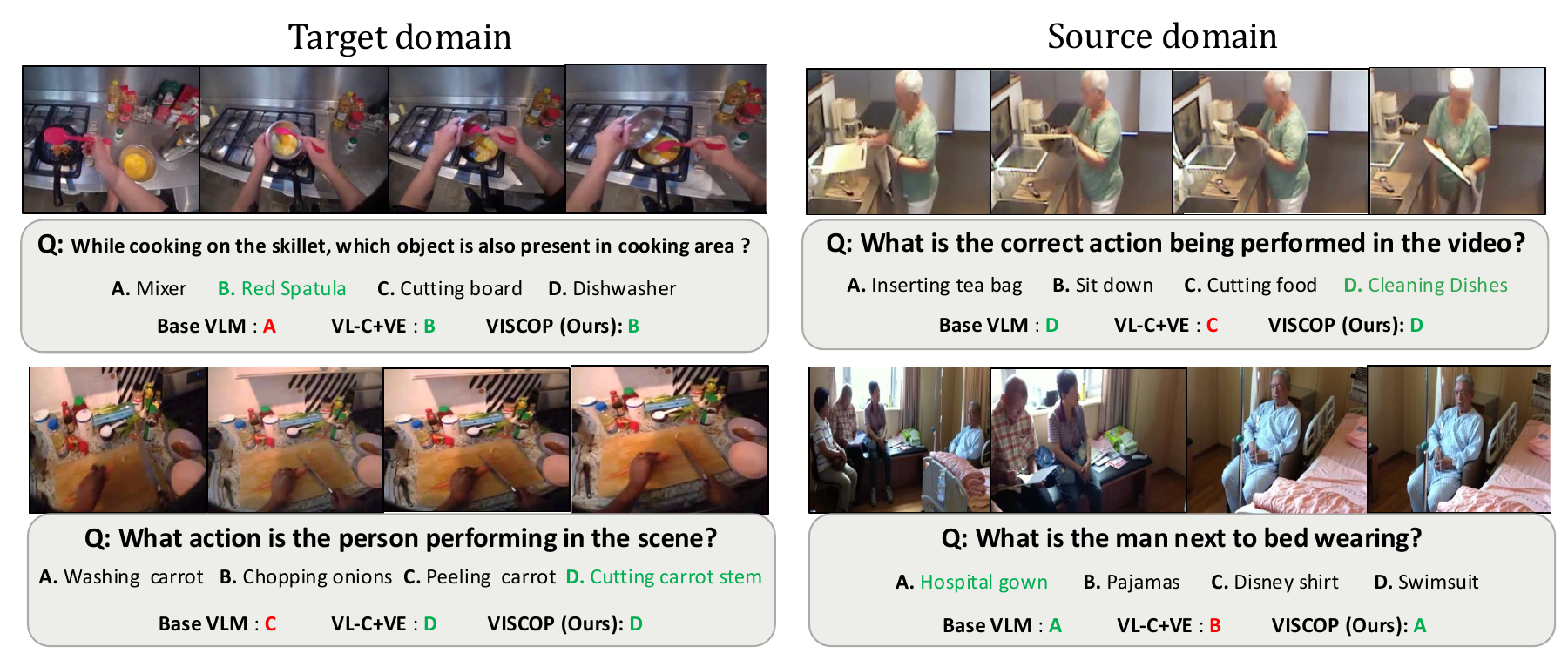}
    \caption{\textbf{Qualitative results on Egocentric Video Understanding Experts.}}
    \label{fig:ego_qualitative}
\end{figure*}

\begin{figure*}[t]
    \centering
    \includegraphics[width=\textwidth]{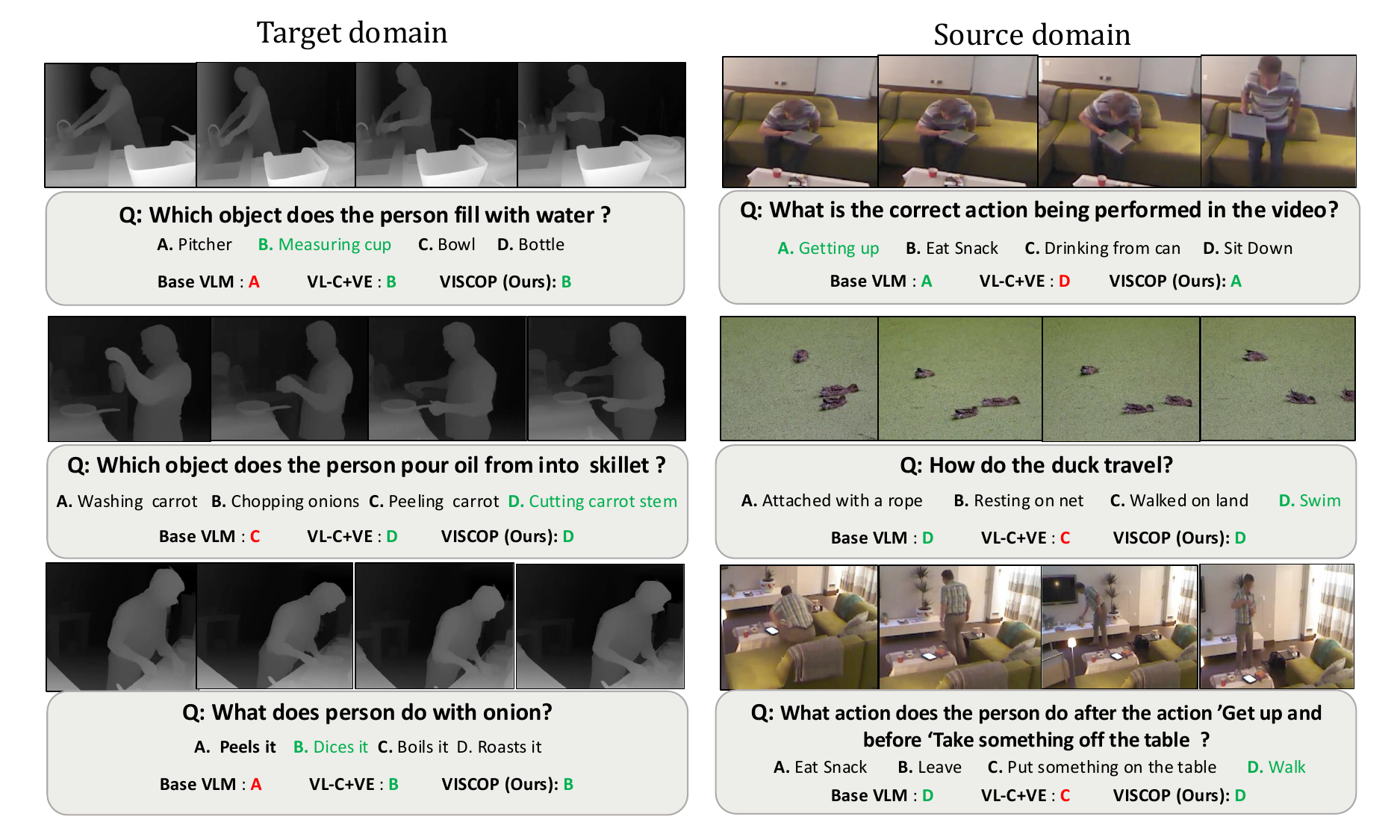}
    \caption{\textbf{Qualitative results on Depth Video Understanding Experts.}}
    \label{fig:depth_qualitative}
\end{figure*}

\begin{figure*}[t]
    \centering
    \includegraphics[width=\textwidth]{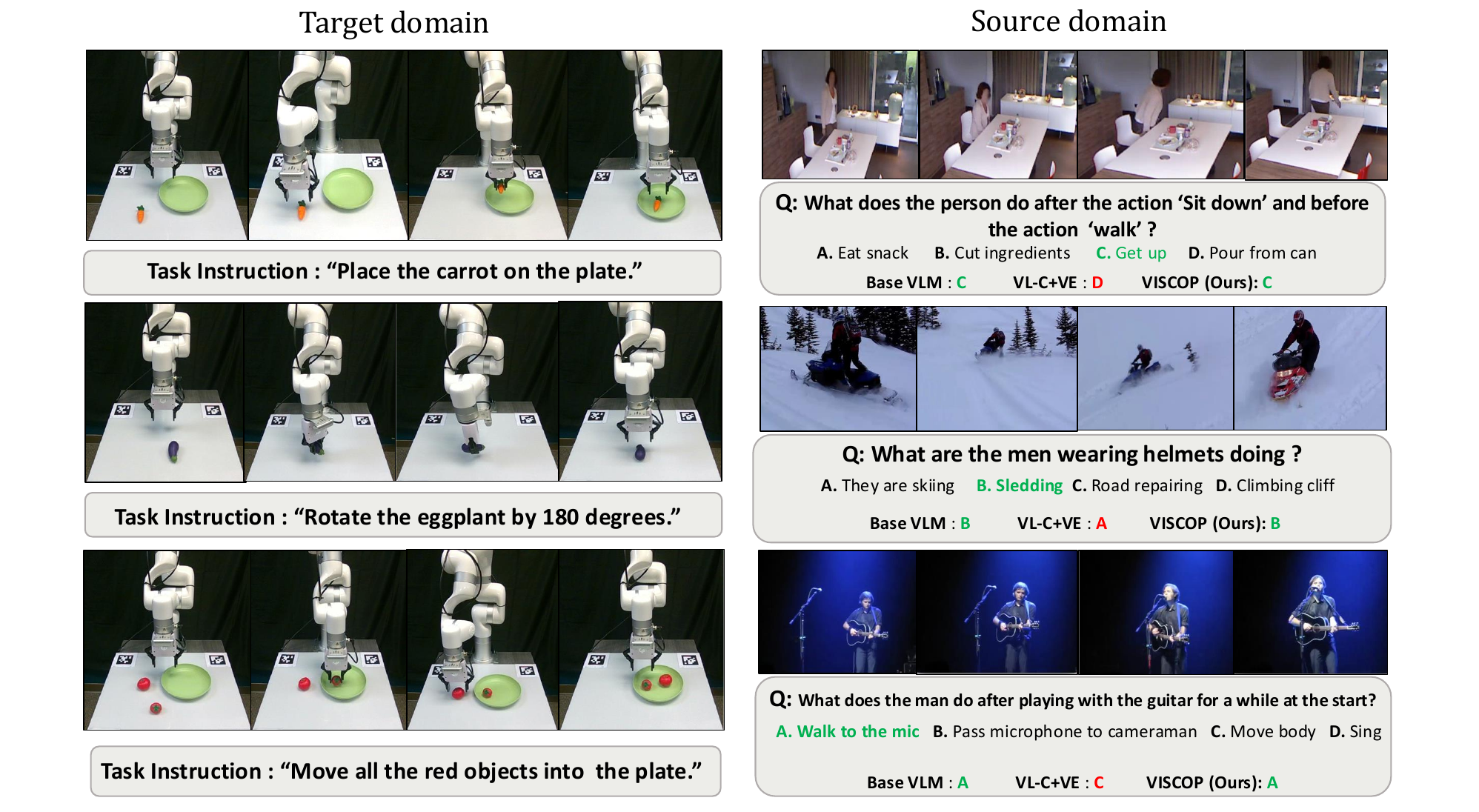}
    \caption{\textbf{Qualitative results on robot control Experts.}}
    \label{fig:robot_qualitative}
\end{figure*}

In this section, we provide qualitative comparisons of three models—Base VLM, trained vision encoder (VL-C+VE), and \methodname~ across the three domain experts  ego-video understanding, depth-video understanding, and robot control. Figures~\ref{fig:ego_qualitative}, \ref{fig:depth_qualitative}, and \ref{fig:robot_qualitative} show representative examples from each expert. 
 Each figure shows representative samples from both the target domain and the source domain.

We demonstrate that  VL-C+VE successfully adapts the Base VLM to the target domain, enabling correct predictions. However, this adaptation comes at the expense of source-domain performance, where VL-C+VE frequently makes mistakes. In contrast, \methodname~ achieves the best of both: it adapts effectively to the target domain while simultaneously retaining strong performance on the source domain, thereby avoiding catastrophic forgetting.

\begin{figure*}[t]
    \centering
    \includegraphics[width=\textwidth]{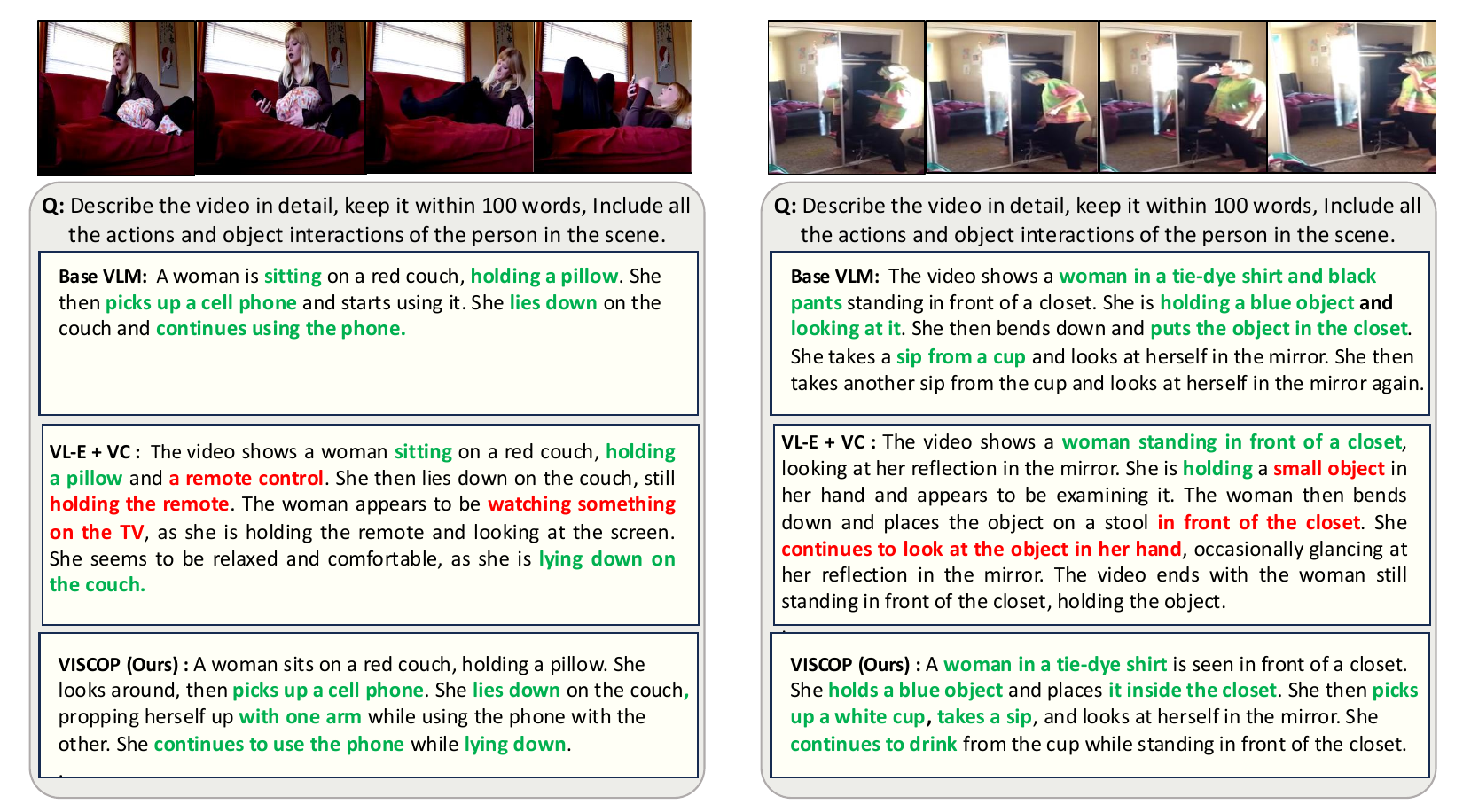}
    \caption{  \textbf{ADL-X descriptions using Ego Video Understanding Expert.}}
    \label{fig:ego_description}
\end{figure*}

\begin{figure*}[t]
    \centering
    \includegraphics[width=\textwidth]{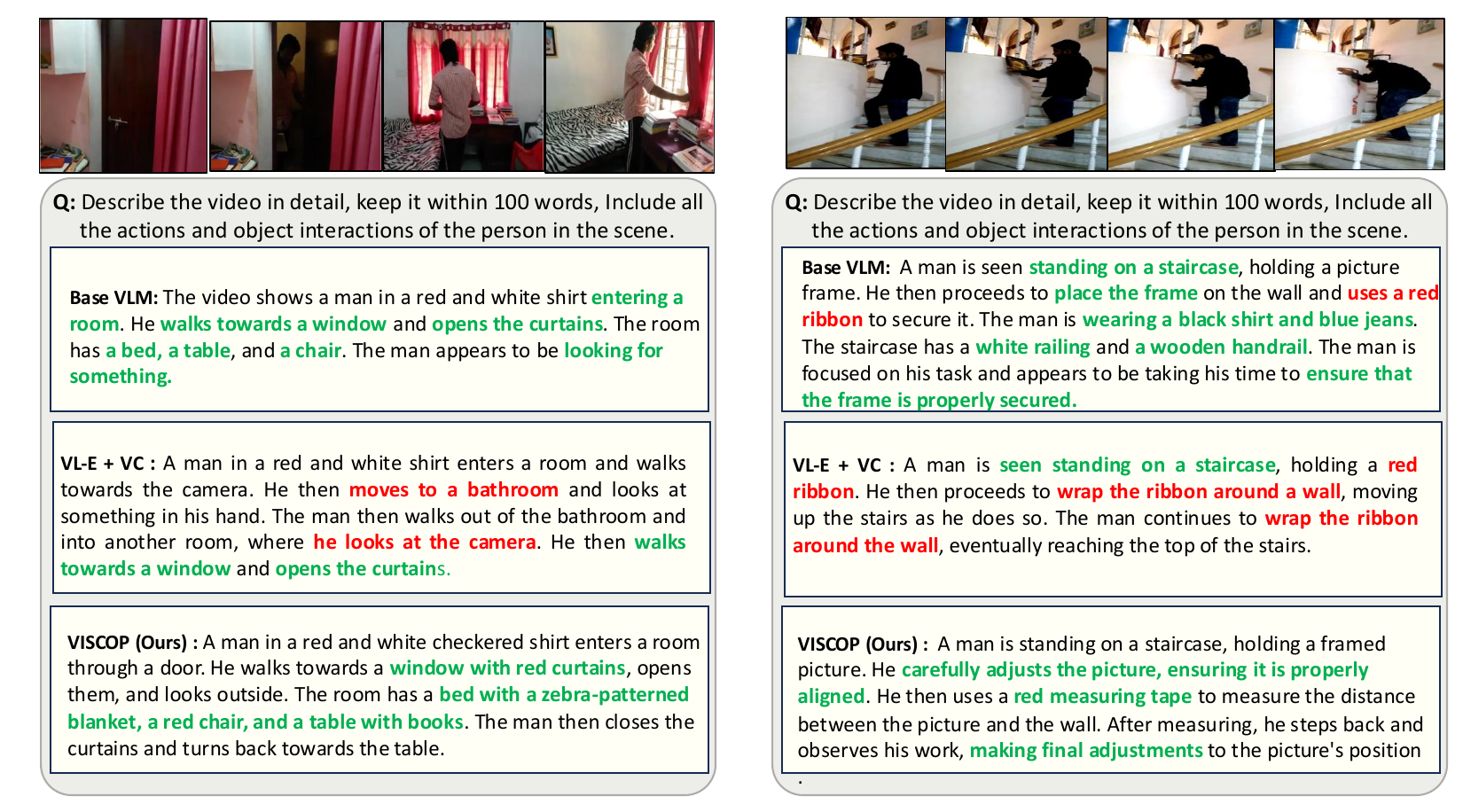}
    \caption{\textbf{ADL-X captions from  the Depth Video Understanding Expert}.}
    \label{fig:depth_description}
\end{figure*}

We also provide qualitative comparisons of video descriptions on the source domain (ADL-X) using the ego-video understanding expert and the depth-video understanding expert. As shown in Figure~\ref{fig:ego_description} and Figure~\ref{fig:depth_description}, our method  generates descriptions that are both more accurate and more detail-oriented compared to the trained vision encoder (VL-C+VE). While VL-C+VE can adapt to the target domain, on the source domain it often introduces hallucinated details. In contrast, \methodname~ preserves correctness, capturing the scene,  actions and object interactions without hallucination.

\end{document}